\ifdefined\pdfoutput
\pdfoutput=1
\fi
\PassOptionsToPackage{numbers,sort&compress}{natbib}
\documentclass{article}

\usepackage[final, main]{neurips_2026}

\usepackage{ifluatex}
\ifluatex
\usepackage[T1]{fontenc}
\fi

\usepackage{microtype}

\renewcommand{\paragraph}[1]{\noindent\textbf{#1}}

\usepackage[ruled,vlined,linesnumbered]{algorithm2e}
\usepackage[labelformat=simple]{subcaption}
\usepackage{graphicx}
\usepackage{wrapfig}
\usepackage{placeins}

\usepackage{amsmath}
\usepackage{amssymb}
\usepackage{amsthm}

\usepackage{enumitem}

\newtheorem{claim}{Claim}

\usepackage[dvipsnames,table]{xcolor}
\colorlet{highlight}{BurntOrange!15}

\usepackage{booktabs}
\usepackage{tabularray}[2022/11/01]
\UseTblrLibrary{booktabs}
\SetTblrInner{
  columns = {colsep=4pt},
  rows= {rowsep=.25pt}
}

\usepackage{xspace}
\makeatletter
\DeclareRobustCommand\onedot{\futurelet\@let@token\@onedot}
\def\@onedot{\ifx\@let@token.\else.\null\fi\xspace}

\def\eg{{e.g}\onedot}
\def\ie{{i.e}\onedot}
\def\cf{{cf}\onedot}
\makeatother

\usepackage{hyperref}

\usepackage[capitalize]{cleveref}
\Crefname{figure}{Fig.}{Figs.}
\Crefname{claim}{Claim}{Claims}

\hypersetup{
  breaklinks,
  colorlinks,
  linkcolor = BrickRed,
  citecolor = RoyalBlue,
  urlcolor  = WildStrawberry,
}

\newcommand\nomarkfootnote[1]{\begingroup
  \renewcommand\thefootnote{}\footnote{#1}\addtocounter{footnote}{-1}\endgroup
}

\makeatletter
\DeclareRobustCommand{\METHODNAME}{PMT\texorpdfstring{\@ifnextchar.{\@}{\xspace}}{}}
\DeclareRobustCommand{\METHODFULLNAME}{Progressive Memory Transformer\texorpdfstring{\xspace}{}}
\makeatother

\newcommand{\sd}[1]{\textsubscript{$\pm$#1}}
\newcommand{\cls}{\texttt{[CLS]}\xspace}

\author{
  \begin{tabular}{@{}c@{\qquad}c@{}}
    Tord Sture Stangeland$^{1,2}$ & Andreas K\"ohler$^{2,3}$ \\
    \normalfont\small\texttt{tord.stangeland@norsar.no} & \normalfont\small\texttt{andreas.kohler@norsar.no} \\[4pt]
    Steffen M\ae land$^{2,4}$ & Ad\'in Ram\'irez Rivera$^1$ \\
    \normalfont\small\texttt{steffen.meland@hvl.no} & \normalfont\small\texttt{adinr@uio.no}
  \end{tabular} \\[6pt]
  \normalfont\small $^1$University of Oslo, Department of Informatics, Oslo, Norway \\
  \normalfont\small $^2$NORSAR, Kjeller, Norway \\
  \normalfont\small $^3$UiT The Arctic University of Norway, Department of Geosciences, Troms\o, Norway \\
  \normalfont\small $^4$Western Norway University of Applied Sciences, Bergen, Norway
}

\title{Progressive Memory Transformer: Memory-Aware Attention for Time-Series}

\begin{document}
\maketitle

\nomarkfootnote{Code: \url{https://github.com/dsb-ifi/pmt}}

\begin{abstract}
Time-series carry structure simultaneously at multiple scales (fine-grained variation, mid-range motifs, and global properties) and downstream tasks operate at correspondingly different scales.
Most existing self-supervised learning approaches supervise representations globally via instance-level contrastive losses and limited temporal neighborhood supervision, but do not explicitly exploit the structural hierarchy.
We propose a learning framework that explicitly enforces a structural hierarchy across three scales independently: a local objective for token alignment, a mid-range objective for window-level motifs, and a global objective for sequence-level agreement.
Realizing this framework requires the backbone to expose a representation at each scale; we introduce \textbf{Progressive Memory Transformer} (PMT), which augments a transformer with writable, window-aligned memory that exposes the mid-range scale alongside the token and sequence-level representations conventional transformers already provide.
Across eight UCR/UEA/UCI classification benchmarks, a cue-retention probe, and forecasting benchmarks, PMT learns representations that probe well at the global, mid-range, and local scales---strong low-label classification (1--5\% labels), competitive forecasting performance across multiple horizons, and quantitative and qualitative evidence that memory states capture mid-range motifs.

\end{abstract}

\section{Introduction}\label{sec:introduction}

Time-series carry structure simultaneously at multiple scales: fine-grained variation within short windows, recurring patterns at intermediate ranges, and global properties of the whole sequence.
Representations that are useful for downstream tasks must capture this structure at every scale where it exists, since downstream tasks (such as local forecasting, motif detection, and sequence-level classification) operate at correspondingly different scales.

\begin{figure}[tb]
\centering

\includegraphics{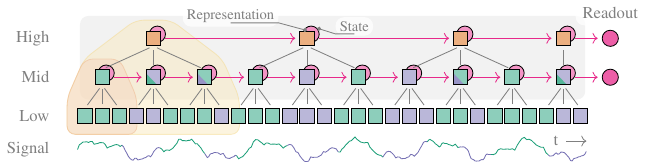}
\caption{An illustration of the three levels of representation and how they capture distinct structural patterns. Shaded regions indicate expanding receptive fields. Exposing the intermediate representations (gray background) enables direct supervision at multiple scales, unlike standard global readouts.}
\label{fig:tree-hierarchy}
\end{figure}

Learning these multi-scale representations is not trivial.
Conventional self-supervised learning (SSL) for time-series typically relies on global instance-level contrastive losses and limited temporal neighborhood supervision~\cite{oord2018representation, chen2020simple, ijcai2021-324, yue2022ts2vecuniversalrepresentationtime, lee2024soft}---at the readout states shown in \Cref{fig:tree-hierarchy}.
Although these methods provide some temporal alignment, they do not explicitly exploit the underlying structural hierarchy of the signal.
Consequently, the intermediate scale (where much of the meaningful temporal structure lives) is often shaped indirectly or destructively compressed (\eg, via max-pooling), rather than explicitly targeted through a dedicated representation interface---shown in gray in \Cref{fig:tree-hierarchy}.
Even architectures designed to propagate context across time (\eg, recurrent hidden states or read-only segment caches) treat their intermediate state as an internal optimization device rather than as a representation exposed to direct, scale-appropriate supervision.
Furthermore, while recent memory-augmented transformers introduce persistent writable slots, they typically operate as global latent banks decoupled from any local temporal alignment, making them unsuitable for targeted mid-range supervision.

To address this, we propose a learning framework that explicitly enforces a structural hierarchy across three scales independently: token-level (fine-grained), mid-range (intermediate motifs), and sequence-level (global)---\cf \Cref{fig:tree-hierarchy}.
Realizing this framework requires the backbone to expose a representation at each scale. To this end, we introduce the \textbf{\METHODFULLNAME} (\METHODNAME), which augments a transformer with a \emph{sample-specific}, writable, window-aligned memory that exposes the mid-range scale alongside the token and sequence-level outputs that conventional transformers already provide.
We use contrastive objectives at each scale, chosen to match the granularity of the representation: local alignment at the token level, mid-range motif consistency at the memory level, and sequence-level agreement at the \cls level.

\paragraph{Contributions.}
(1)~We propose a multi-scale contrastive learning framework for time-series that supervises representation at three scales independently: a token-level Hierarchical Gaussian Contrastive Loss (\emph{HGCL}), a mid-range memory-state loss (\emph{PCL}), and a sequence-level instance loss (\emph{ICL}).
(2)~We introduce \METHODNAME, a memory-augmented transformer backbone whose \emph{window-aligned writable memory} exposes the mid-range scale as a directly supervisable representation, complementing the token and sequence-level outputs of conventional transformers.
(3)~We evaluate the effectiveness of the resulting representations at three scales: forecasting on standard benchmarks for the local level, cue retention plus qualitative analysis for the mid-range level, and low-label classification (1--5\% labels) on eight UCR/UEA/UCI benchmarks for the global level, alongside fully supervised validations (\Cref{app:supervised}) demonstrating its utility across different tasks.

\FloatBarrier
\section{Progressive Memory Transformers}
\label{sec:method}

Standard time-series architectures compress temporal dynamics into a single global vector, treating intermediate states as transient variables that are aggregated away.
By simply discarding these hidden states (shown in the gray background of \Cref{fig:tree-hierarchy}), existing methods lose the distinct recurring motifs that emerge at intermediate scales (denoted by the shaded patterns on the representations in the figure).
This section instantiates our alternative: a multi-scale framework that explicitly exposes and supervises representations at three distinct resolutions---fine-grained tokens, mid-range motifs, and global summaries.
While conventional transformers already expose token and sequence-level outputs, the intermediate representations remain hidden.
We introduce \textbf{Progressive Memory Transformer (PMT)}, which augments a transformer with writable, window-aligned memory states that expose the mid-range scale as a first-class representation.

By keeping all three levels accessible, we can ground them with scale-matched contrastive objectives to extract useful information at the exact granularity downstream tasks require.
We first introduce \textbf{Progressive Memory Attention (PMA)} (\Cref{sec:pma}), the architectural block that produces these mid-range memory states alongside refined tokens.
Because these memory states are explicitly exposed, we supervise them directly with the \textbf{PMA Contrastive Loss (PCL)} (\Cref{sec:pcl}) to capture robust regional structure.
This mid-range objective is complemented by token- and sequence-level contrastive objectives (\Cref{sec:losses}), and the full pipeline is composed in \Cref{sec:pipeline}.

\paragraph{Notation.}\label{pag:notation}
Let $x \in \mathbb{R}^{L \times C}$ denote a single $C$-channel time-series of length $L$.
The encoder $f_\theta$ produces a matrix $R \in \mathbb{R}^{K \times D}$ containing $K$ \emph{tokens}, each $D$-dimensional, plus a separate \cls vector $c \in \mathbb{R}^{D}$.
Hence, the overall output is $\bigl[R;\, c\bigr] \in \mathbb{R}^{(K+1) \times D}$.

\subsection{Progressive Memory Attention}
\label{sec:pma}

\begin{wrapfigure}{R}{0.35\textwidth}\centering
  \resizebox{\linewidth}{!}{\includegraphics{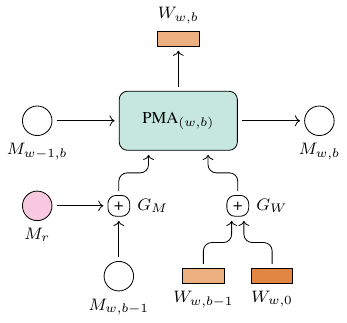}}
\caption{PMA information flow~\eqref{eq:pma}.}
\label{fig:pma}
\end{wrapfigure}

The core engine of our architecture is the \textbf{Progressive Memory Attention (PMA)} operation, illustrated in \Cref{fig:pma}.
Instead of independently processing isolated windows or maintaining a strictly read-only cache, PMA equips each window with a small, fixed-size set of writable memory slots.
For a given window $w$ at block $b$, the PMA operation takes the window's tokens alongside $n_m$ memory slots carried forward from the previous window, producing a refined token stream $W_{w,b}$ and an updated memory state $M_{w,b}$
\begin{equation}
    \label{eq:pma}
    W_{w,b},\, M_{w,b} \;=\; \mathrm{PMA}\!\left(M_{w-1,b},\, \bar{M}_{w,b-1},\, \bar{W}_{w,b-1}\right).
\end{equation}
These inputs encode context along two axes: $M_{w-1,b}$ propagates memory \textbf{horizontally} from the previous window at the same block, while $\bar{M}_{w,b-1}$ and $\bar{W}_{w,b-1}$ carry the refined memory and tokens \textbf{vertically} from the previous block to the next level.
(The bar denotes variables refined by adaptive gates.)
Rather than forcing the model to re-attend to all past tokens, this carried memory acts as a compact, evolving summary of historical context.

Concretely, for each window PMA first refines the previous-block memory by mixing it with a learned reset state, concatenates this refined memory with the carried memory and the current window tokens, applies an asymmetric attention mask that keeps memory updates strictly causal, and reads out the updated memory slots alongside the refined window tokens (\Cref{app:pma_forward} gives the algorithmic form).
A single multi-head attention layer is applied jointly to the concatenated sequence; because the slots serve as both queries and keys/values, the attention mechanism can read from and write to them in a single pass.

Two adaptive gates regulate this flow.
The first softly mixes the carried memory with a learned reset state (analogous to the forget/reset gates in recurrent architectures), allowing the model to overwrite stale context when temporal regimes change.
The second mixes the processed window tokens with their original signal representation, preserving fine-grained local detail while integrating the broader context.
Both gates, the attention-mask construction, and the per-window overlap aggregation are detailed in \Cref{app:pma-impl}.

Crucially, these persistent, window-aligned memory slots serve as the \textbf{interface} through which the mid-range level becomes explicitly supervisable.
Because they are updated in place and exposed at the window scale, we can apply an objective (PCL, \Cref{sec:pcl}) directly to them, rather than indirectly via the sequence summary.

By applying this operation to every window, a \textbf{PMA block} propagates memory horizontally across time.
Stacking $B$ such blocks into the full PMT backbone propagates information vertically, expanding a token's contextual span linearly in both depth and window index.
This dual propagation allows deeper blocks to effectively fuse historical summaries (\Cref{app:rfield} provides a formal receptive-field derivation).

\subsection{PMA Contrastive Loss}
\label{sec:pcl}

To ground these memory representations, we apply the \textbf{PMA Contrastive Loss (PCL)} (an InfoNCE objective) directly to the memory slots emerging from the final PMA block.
By demanding cross-view consistency at the window scale, PCL forces the model to distill salient mid-range motifs into these memory slots such that they are persistent over the stochastic augmentations.
Intuitively, each memory slot $m$ functions as a learnable motif detector for its respective window (an individual $D$-dimensional vector extracted from the final block's memory state $M$; see \Cref{app:pcl-impl} for its formal definition). Contrasting these individual slots directly ensures that they do not simply cache raw past tokens, but actively compress the window's temporal dynamics into distinct, robust semantic features.

Specifically, each anchor memory slot $m_a$ from one view is paired with its corresponding positive slot $m_p$ from the second view at the same window and slot index.
Negatives $m_n \in \mathcal{N}_a$ are drawn from the memory slots of other sequences in the mini-batch.
The loss takes the standard InfoNCE~\citep{oord2018representation} form
\begin{equation}
    \label{eq:pcl}
    \mathcal{L}_\text{PCL} = -\mathbb{E}_{m_a} \log \frac{\exp(\langle m_a, m_p \rangle / \tau)}{\exp(\langle m_a, m_p \rangle / \tau) + \sum_{m_n \in \mathcal{N}_a} \exp(\langle m_a, m_n \rangle / \tau)},
\end{equation}
where $\langle \cdot, \cdot \rangle$ denotes cosine similarity (slots are $\ell_2$-normalized) and $\tau$ is a temperature parameter.
(Full notation, symmetric anchor/key averaging, and implementation details are provided in \Cref{app:pcl-impl}.)

\subsection{Token and Sequence Objectives}
\label{sec:losses}

To complete our multi-scale hierarchy, the local (token) and global (sequence) representation levels are supervised by two complementary contrastive objectives.
Both follow the standard protocol of projecting representations through a level-specific head prior to the loss.

\paragraph{Hierarchical Gaussian Contrastive Loss (HGCL).}
HGCL grounds the fine-grained token representations by matching each token to its temporally aligned counterpart in the second view.
Because tokens at neighboring positions within the same sequence are intrinsically correlated, HGCL does not treat them as individual negatives: their similarities are weighted by a Gaussian over temporal distance and combined into a single term of the objective, while cross-sequence tokens serve as hard negatives.
Crucially, these Gaussian widths are derived directly from the window size and stride, requiring no pre-computed data-space distances or alignments.

This objective operates hierarchically at two scales: a token-level component, $\mathcal{L}_t$, that softly aggregates neighbors within the same window, and a window-level component, $\mathcal{L}_w$, that aggregates overlapping windows across the sequence.
For both, the anchor's positive is its aligned counterpart in the second view, the Gaussian-weighted aggregate of nearby tokens forms a separate term in the denominator, and negatives are drawn from other sequences in the batch.
The total loss is their weighted sum
\begin{equation}
    \label{eq:hgcl}
    \mathcal{L}_\text{HGCL} = \alpha\, \mathcal{L}_{t} + \beta\, \mathcal{L}_{w},
\end{equation}
where $\alpha$ and $\beta$ weight the token and window levels.
(The bucket-form InfoNCE structure and exact Gaussian formulations are detailed in \Cref{app:hgcl-impl}.)

\paragraph{Instance Contrastive Loss (ICL).}
Finally, ICL grounds the global sequence summary using a standard instance-level InfoNCE objective.
For a mini-batch of $\mathcal{B}$ sequences with two stochastic views each, let $c_i^v$ denote the \cls token of the $i$-th sequence under view $v$.
Each anchor $g(c_i^1)$ pairs with its matched view-$2$ counterpart $g(c_i^2)$ as the positive, while the remaining $2(\mathcal{B}-1)$ projected \cls tokens in the batch serve as negatives
\begin{equation}
    \label{eq:icl}
    \mathcal{L}_\text{ICL} = -\frac{1}{\mathcal{B}} \sum_{i=1}^{\mathcal{B}} \log \frac{\exp(\langle g(c_i^1), g(c_i^2) \rangle / \tau)}{\sum_{(i', v) \neq (i, 1)} \exp(\langle g(c_i^1), g(c_{i'}^v) \rangle / \tau)},
\end{equation}
where $\langle \cdot, \cdot \rangle$ denotes cosine similarity and $g(\cdot)$ is the projection head.
(The encoder that aggregates the PMA token stream into the sequence summary $c$ is described in \Cref{sec:pipeline}; symmetric view averaging and projection-head specifics are in \Cref{app:icl-impl}.)

\subsection{End-to-End Learning Framework}
\label{sec:pipeline}

The full PMT pipeline integrates the components described above into a single, end-to-end forward pass.
(1)~A 1-D convolution tokenizes the $C$-channel waveform into $K$ patch tokens.
(2)~These tokens are unfolded into overlapping windows of length $W$ and stride $S$, and processed by a stack of $B$ PMA blocks where memory propagates both horizontally and vertically (\Cref{sec:pma}).
Because windows overlap ($S < W$), an \emph{attentive overlap aggregator} (\Cref{app:overlap-aggregator}) merges overlapping token outputs after each block to maintain a constant sequence length.
(3)~A short stack of encoder layers refines the resulting token stream using neighborhood-masked attention (restricting each token to a local context window) while a single \cls token attends globally to aggregate the sequence summary.
(4)~The network is trained by explicitly grounding the structural hierarchy at its three representation levels: HGCL on the fine-grained tokens, PCL on the mid-range memory slots, and ICL on the global sequence summary (\Cref{sec:pcl,sec:losses}).
These three exposed levels---local tokens, mid-range memory, and global summary---serve directly as the distinct representation interfaces for downstream tasks.
The total loss is the weighted sum of these three objectives, \ie,
\begin{equation}
    \label{eq:total-loss}
    \mathcal{L}_\text{total} = \lambda_\text{ICL}\mathcal{L}_\text{ICL} + \lambda_\text{HGCL}\mathcal{L}_\text{HGCL} + \lambda_\text{PCL}\mathcal{L}_\text{PCL}.
\end{equation}

\paragraph{Two-view training.}
Following standard practice in time-series contrastive learning~\citep{Eldele_2023, Eldele_2024, yue2022ts2vecuniversalrepresentationtime}, each sample is presented to the weight-shared backbone as two stochastic views.
A \textbf{weak} view preserves local morphology (via channel-wise z-scaling and low-variance Gaussian noise), while a \textbf{strong} view additionally breaks short-range correlations while preserving coarse semantics (via time-warping and magnitude-warping).
The three contrastive objectives operate jointly on these paired views, denoted as view~$1$ and view~$2$.

Full implementation details (including dataset-specific normalization, attention masking specifics, and the per-block handling of overlapping window outputs) are deferred to \Cref{app:implementation}; \Cref{app:compute} reports compute and memory cost relative to a FlashAttention baseline.

\FloatBarrier
\section{Experiments}
\label{sec:experiments}

We evaluate \METHODNAME by probing each of its three representation levels with a downstream task matched to that level's temporal granularity: low-label classification at the global level (\Cref{sec:exp:global}), cue retention and qualitative analysis at the mid-range level (\Cref{sec:exp:midrange}), and forecasting at the local level (\Cref{sec:exp:local}).
We then isolate the architectural contribution from the multi-scale objectives through a matched-architecture comparison (\Cref{sec:exp:arch}) and quantify loss complementarity in \Cref{sec:exp:loss}.

We use eight UCR/UEA/UCI benchmarks~\citep{UCRArchive2018,UEAArchive2018,anguita2013public}: \textit{HAR}, \textit{Epilepsy}, \textit{Wafer}, \textit{FordA}, \textit{FordB}, \textit{Phalanges\-Outlines\-Correct (POC)}, \textit{Electric\-Devices}, and \textit{StarLightCurves}.
For classification we follow the evaluation protocol of \citet{ijcai2021-324}: the backbone is pretrained without labels, frozen, and an SVM classifier with an RBF kernel, as in \citet{lee2024soft}, is trained on 1\% and 5\% labeled subsets; we report top-1 accuracy and macro-F1 (mF1) on the official test split.
Forecasting and cue retention follow the protocols introduced in their respective subsections.
We compare against time-series SSL baselines TS2Vec~\citep{yue2022ts2vecuniversalrepresentationtime}, TS-TCC~\citep{ijcai2021-324}, and SoftCLT~\citep{lee2024soft}, and against generic SSL baselines adapted to time-series: CPC~\citep{oord2018representation}, SimCLR~\citep{chen2020simple}, and SSL-ECG~\citep{Sarkar_2020}.
Dataset selection, per-dataset statistics, and additional protocol details are in \Cref{app:protocol}; related work appears in \Cref{sec:related_work}.

\subsection{Global Level} \label{sec:exp:global}

\begin{table}[tb]
  \centering
  \footnotesize
  \caption{Self-supervised learning results on $1\%$ and $5\%$ labeled subsets.
  Each cell shows accuracy / macro-F1. For \METHODNAME, each value is the mean over ten pretraining seeds, with the sample standard deviation as a subscript.}
  \label{tab:classification_table}
\footnotesize
\begin{tblr}{
  colspec={l*{6}{r}},
  colsep=4pt,
  column{7}={bg=highlight},
  row{1,2,11,12}={bg=white, rowsep=2pt},
}
    \toprule
    & \SetCell[c=6]{c}{Self-supervised learning (1\% labeled)} &&&&& \\
    \cmidrule[lr]{2-7}
    Dataset & SSL-ECG & CPC & SimCLR & TS2Vec+SoftCLT & TS-TCC+SoftCLT & \METHODNAME \\
    \midrule
    HAR & 60.0 / 54.0 & 65.4 / 63.8 & 65.8 / 64.3 & \underline{91.0} / \underline{91.0} & 82.9 / 82.8 & \textbf{92.0}\sd{1.4} / \textbf{92.0}\sd{1.4} \\
    Epilepsy & 89.3 / 86.0 & 88.9 / 85.8 & 88.3 / 84.0 & \underline{96.3} / 94.1 & 95.6 / \textbf{95.6} & \textbf{96.6}\sd{0.2} / \underline{94.7}\sd{0.3} \\
    Wafer & 93.4 / 76.1 & 93.5 / 78.4 & 93.8 / 78.5 & 95.3 / 88.1 & \underline{96.5} / \textbf{96.5} & \textbf{98.0}\sd{0.9} / \underline{94.6}\sd{2.8} \\
    FordA & 67.9 / 66.2 & 75.8 / 75.2 & 55.9 / 55.7 & \underline{87.1} / \underline{87.1} & 81.5 / 81.2 & \textbf{87.8}\sd{2.1} / \textbf{87.7}\sd{2.1} \\
    FordB & 64.4 / 60.5 & 66.8 / 65.0 & 50.9 / 49.8 & 67.9 / 67.9 & \underline{74.8} / \underline{74.8} & \textbf{75.6}\sd{1.9} / \textbf{75.3}\sd{2.3} \\
    POC & 62.5 / 41.2 & 64.8 / 48.2 & 61.5 / 38.4 & 63.6 / \underline{62.8} & \underline{65.4} / \textbf{64.6} & \textbf{65.9}\sd{3.2} / 61.1\sd{4.7} \\
    ElectricDevices & 60.1 / 50.0 & 59.3 / 48.9 & \underline{62.5} / 51.2 & 62.0 / \underline{53.0} & \textbf{64.6} / \textbf{63.2} & 60.2\sd{3.0} / \underline{53.0}\sd{5.2} \\
    StarLightCurves & 78.3 / 72.0 & 80.8 / 74.4 & 80.6 / 71.6 & 85.6 / 62.9 & \underline{86.0} / \underline{79.3} & \textbf{92.6}\sd{3.2} / \textbf{88.4}\sd{9.3} \\
    \midrule
    & \SetCell[c=6]{c}{Self-supervised learning (5\% labeled)} &&&&& \\
    \cmidrule[lr]{2-7}
    Dataset & SSL-ECG & CPC & SimCLR & TS2Vec+SoftCLT & TS-TCC+SoftCLT & \METHODNAME \\
    \midrule
    HAR & 63.7 / 58.6 & 75.4 / 74.7 & 75.8 / 74.9 & 92.1 / 92.1 & \underline{92.6} / \underline{92.6} & \textbf{93.4}\sd{0.7} / \textbf{93.5}\sd{0.7} \\
    Epilepsy & 92.8 / 89.0 & 92.8 / 90.2 & 91.3 / 89.2 & \underline{96.7} / 94.9 & 96.2 / \textbf{96.1} & \textbf{96.9}\sd{0.2} / \underline{95.1}\sd{0.3} \\
    Wafer & 94.9 / 84.5 & 92.5 / 79.4 & 94.8 / 83.3 & \underline{98.8} / \underline{96.8} & 98.2 / \textbf{98.2} & \textbf{98.8}\sd{0.2} / 96.7\sd{0.4} \\
    FordA & 73.6 / 70.7 & 86.5 / 86.5 & 69.6 / 68.9 & \underline{92.5} / \underline{92.5} & \textbf{93.2} / \textbf{93.2} & 89.5\sd{0.5} / 89.5\sd{0.5} \\
    FordB & 71.7 / 69.8 & \underline{86.3} / \underline{86.2} & 63.0 / 60.7 & 78.8 / 78.6 & \textbf{88.0} / \textbf{88.0} & 76.9\sd{0.5} / 76.9\sd{0.5} \\
    POC & 62.9 / 43.3 & 66.9 / 44.3 & 62.7 / 42.4 & \underline{70.9} / \underline{69.7} & 69.4 / 66.3 & \textbf{72.7}\sd{2.8} / \textbf{71.4}\sd{2.6} \\
    ElectricDevices & 63.7 / 56.1 & 62.4 / 58.1 & 63.9 / 58.6 & 62.4 / 54.4 & \underline{65.1} / \textbf{63.8} & \textbf{65.9}\sd{1.7} / \underline{60.3}\sd{2.3} \\
    StarLightCurves & 82.6 / 74.5 & 89.1 / 84.5 & 84.2 / 74.0 & \underline{92.3} / \underline{89.8} & 86.2 / 85.5 & \textbf{96.9}\sd{0.6} / \textbf{95.5}\sd{0.7} \\
    \bottomrule
\end{tblr}
\end{table}

\Cref{tab:classification_table} presents classification results at 1\% and 5\% labels.
On average across all eight datasets and both label fractions, \METHODNAME achieves the highest Top-1 accuracy ($85.0$ vs.\ $83.5$ for TS-TCC+SoftCLT and $83.3$ for TS2Vec+SoftCLT) and the highest macro-F1 score, outperforming or matching baselines in 13 out of 16 accuracy settings.
The largest margins are on StarLightCurves, the longest sequences in the suite ($+6.6$ and $+4.6$pp Top-1 at 1\% and 5\%).

The engine-vibration benchmarks (FordA/B) present an exception.
While \METHODNAME leads at 1\% labels on both, its performance stagnates at 5\% labels.
In contrast, autoregressive and predictive-contrastive baselines (CPC, TS-TCC+SoftCLT) gain $11$--$20$pp and the purely contrastive multi-resolution baseline TS2Vec+SoftCLT gains $5$--$11$pp, whereas \METHODNAME remains close to its 1\% accuracy.
We attribute this to two likely factors: first, the short-lived phase and frequency events characteristic of FordA/B are highly sensitive to the tokenizer's stride (\Cref{sec:patch-and-stride-ab}); second, predictive pretraining naturally encodes phase-relative temporal positions, giving it an advantage over the agreement-based contrastive objectives used by \METHODNAME.
Integrating a predictive component to bridge this gap remains outside of the scope of this work.

\subsection{Mid Range} \label{sec:exp:midrange}

The mid-range probe tests whether the writable memory states function as a representation level rather than an internal optimization device. We design two evaluations: a quantitative cue-retention probe that isolates whether mid-range information is recoverable from the memory states, and a qualitative similarity analysis that inspects what the memory states encode.

\paragraph{Cue retention.}
We design a new cue-retention experiment to test whether a backbone with mid-range state actually \emph{uses} that state to carry information through a sequence.
Strong downstream accuracy does not by itself indicate that an architecture is propagating mid-range information---the task may simply be solvable from local features.
To disentangle the confoundment, our protocol injects a synthetic perturbation (a Hann-tapered burst) at varying distances from the end of the sequence into a frozen backbone that was pretrained on FordA without ever seeing the cue.
A logistic-regression probe then predicts cue presence strictly from end-of-sequence (EOS) features.
Because the model was never trained to detect the cue, success requires that intermediate evidence was actively preserved through state propagation rather than discarded.
The EOS feature is the final memory state for \METHODNAME, the corresponding recurrent or cache state for similar-architecture baselines (\Cref{sec:exp:arch}), and windowized embeddings for TS2Vec.
\Cref{app:retention} details the full protocol.

\begin{figure}[tb]
\centering
\resizebox{0.68\linewidth}{!}{\includegraphics{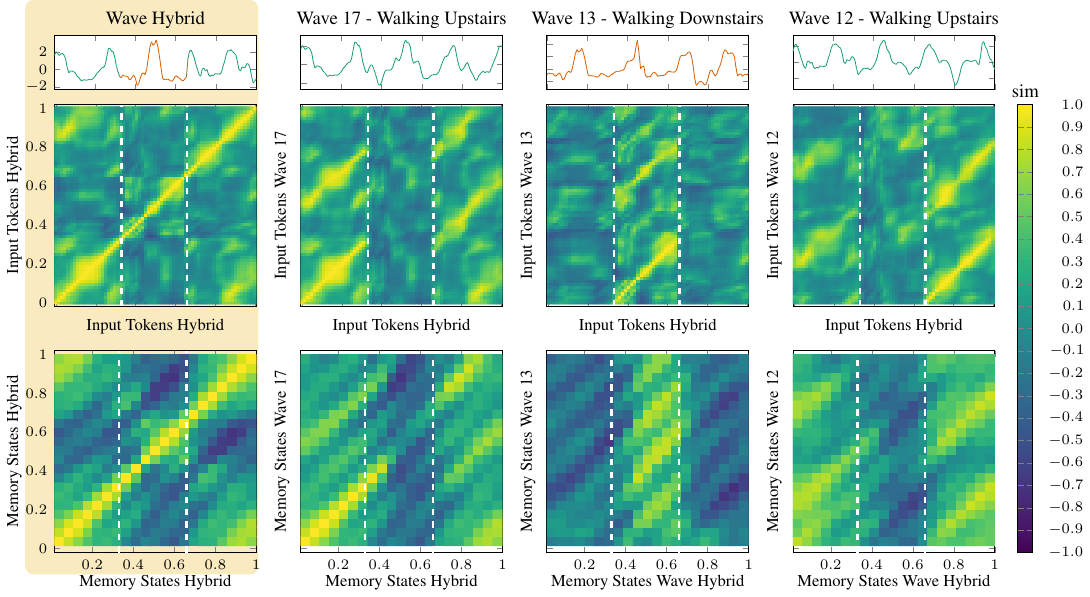}}
\caption{Cosine similarity matrices for the representations and states between pair-wise signals from HAR\@.  Higher similarity shows that the signals correlate as evidenced by the learned embeddings. The vertical bars denote the different sections of the hybrid wave. The wave number corresponds to the sample index in the validation dataset.}
\label{fig:hybrid-sims}
\end{figure}

\begin{wraptable}{R}{0.50\textwidth}
\caption{FordA cue detection. EOS-probe averages across five seeds; subscripts give the sample standard deviation.}
\label{tab:forda-cue-eos}
\centering
\footnotesize
\begin{tblr}{colspec={l*{2}{r}}}
\toprule
Method  & EOS AUC & EOS Top1 \\
\midrule
TS2Vec & 0.639\sd{0.020} & 0.620\sd{0.015} \\
Transformer-XL & 0.967\sd{0.009} & 0.905\sd{0.016} \\
xLSTM & 0.986\sd{0.003} & 0.946\sd{0.011} \\
\METHODNAME-$\lambda_{\text{PCL}}=0$ & 0.962\sd{0.031} & 0.904\sd{0.043} \\
\METHODNAME & \textbf{0.995}\sd{0.002} & \textbf{0.966}\sd{0.008} \\
\bottomrule
\end{tblr}
\end{wraptable}

To apply this protocol to \METHODNAME specifically, we further isolate the writable-memory \emph{interface} from the supervision that shapes it.
We evaluated an unsupervised-memory variant (\METHODNAME-$\lambda_{\text{PCL}}{=}0$, identical architecture but PCL disabled during pretraining).
The $\lambda_{\text{PCL}}{=}0$ row measures the interface alone, and the gap to full \METHODNAME recovers PCL's contribution.

TS2Vec, which relies on hierarchical pooling, discards the mid-range pattern and retains the cue poorly ($0.639$ EOS AUC in \Cref{tab:forda-cue-eos}).
State-propagating architectures fare much better ($0.967$ and $0.986$ AUC for Transformer-XL and xLSTM), and \METHODNAME retains the cue best ($0.995$).
This margin comes from PCL: supervising the memory raises AUC by $+0.033$ and Top-1 by $+0.062$, in all five seeds.

\paragraph{Qualitative analysis.} \Cref{fig:hybrid-sims} shows cosine-similarity matrices on a HAR sequence spliced from three activities (walking upstairs, walking downstairs, walking upstairs). The block-diagonal pattern across same-class segments is sharper at the memory-state level than at the input-token level, with the two upstairs segments mutually bright while the downstairs segment is suppressed. Additional splices and Epilepsy comparisons are in \Cref{app:repr}. Compared to TS2Vec+SoftCLT's intermediate convolutional embeddings, \Cref{fig:softclt_fig4}, PMT's memory states show sharper class-block structure on the same splice.

\subsection{Local Level}
\label{sec:exp:local}

\begin{table}[b]
\caption{Forecasting results following SoftCLT evaluation for different horizons (H); means over five seeds, subscripts give the sample standard deviation.}
\label{tab:forecasting}
\centering
\footnotesize
\resizebox{\linewidth}{!}{\begin{tblr}{
  colsep=3pt,
  colspec={l l *{8}{r}},
  cell{9,15}{2-10}={bg=gray!11},
}
\toprule
 & & \SetCell[c=4]{c} Univariate &&&& \SetCell[c=4]{c} Multivariate &&& \\
\cmidrule[l, r]{3-6}\cmidrule[l, r]{7-10}
 &  & \SetCell[c=2]{c} SoftCLT && \SetCell[c=2]{c} PMT && \SetCell[c=2]{c} SoftCLT && \SetCell[c=2]{c} PMT & \\
\cmidrule[l, r]{3-4}\cmidrule[l, r]{5-6}\cmidrule[l, r]{7-8}\cmidrule[l, r]{9-10}
 Dataset & H & MSE & MAE & MSE & MAE & MSE & MAE & MSE & MAE \\
\midrule
ETTh1 & 24 & \textbf{0.045}\sd{0.004} & \textbf{0.160}\sd{0.007} & 0.054\sd{0.005} & 0.179\sd{0.007} & 0.586\sd{0.027} & 0.530\sd{0.016} & \textbf{0.573}\sd{0.021} & \textbf{0.516}\sd{0.010} \\
 & 48 & \textbf{0.073}\sd{0.006} & \textbf{0.204}\sd{0.009} & 0.075\sd{0.005} & 0.214\sd{0.009} & 0.615\sd{0.026} & 0.552\sd{0.016} & \textbf{0.604}\sd{0.020} & \textbf{0.542}\sd{0.010} \\
 & 168 & 0.152\sd{0.011} & 0.303\sd{0.014} & \textbf{0.133}\sd{0.011} & \textbf{0.285}\sd{0.015} & \textbf{0.753}\sd{0.032} & 0.638\sd{0.018} & 0.755\sd{0.027} & \textbf{0.635}\sd{0.014} \\
 & 336 & 0.170\sd{0.016} & 0.325\sd{0.017} & \textbf{0.150}\sd{0.012} & \textbf{0.308}\sd{0.015} & \textbf{0.899}\sd{0.034} & \textbf{0.717}\sd{0.017} & 0.949\sd{0.041} & 0.734\sd{0.019} \\
 & 720 & 0.333\sd{0.077} & 0.503\sd{0.073} & \textbf{0.157}\sd{0.010} & \textbf{0.324}\sd{0.013} & \textbf{1.016}\sd{0.047} & \textbf{0.785}\sd{0.017} & 1.126\sd{0.063} & 0.824\sd{0.021} \\
 & \textbf{Avg.} & 0.155\sd{0.017} & 0.299\sd{0.016} & \textbf{0.114}\sd{0.006} & \textbf{0.262}\sd{0.007} & \textbf{0.774}\sd{0.032} & \textbf{0.644}\sd{0.016} & 0.801\sd{0.030} & 0.650\sd{0.012} \\
\midrule
Electricity & 24 & 0.260\sd{0.005} & 0.290\sd{0.006} & \textbf{0.253}\sd{0.004} & \textbf{0.272}\sd{0.005} & 0.442\sd{0.018} & 0.481\sd{0.010} & \textbf{0.364}\sd{0.010} & \textbf{0.434}\sd{0.006} \\
 & 48 & 0.308\sd{0.005} & 0.317\sd{0.004} & \textbf{0.303}\sd{0.005} & \textbf{0.305}\sd{0.005} & 0.449\sd{0.018} & 0.484\sd{0.010} & \textbf{0.384}\sd{0.010} & \textbf{0.446}\sd{0.006} \\
 & 168 & 0.426\sd{0.009} & 0.392\sd{0.005} & \textbf{0.415}\sd{0.003} & \textbf{0.377}\sd{0.004} & 0.456\sd{0.015} & 0.489\sd{0.009} & \textbf{0.407}\sd{0.006} & \textbf{0.458}\sd{0.005} \\
 & 336 & 0.566\sd{0.011} & 0.474\sd{0.006} & \textbf{0.555}\sd{0.007} & \textbf{0.456}\sd{0.002} & 0.458\sd{0.015} & 0.492\sd{0.010} & \textbf{0.415}\sd{0.007} & \textbf{0.465}\sd{0.006} \\
 & 720 & 0.863\sd{0.004} & 0.648\sd{0.007} & \textbf{0.857}\sd{0.005} & \textbf{0.638}\sd{0.004} & 0.456\sd{0.013} & 0.495\sd{0.011} & \textbf{0.425}\sd{0.009} & \textbf{0.475}\sd{0.008} \\
 & \textbf{Avg.} & 0.485\sd{0.006} & 0.424\sd{0.003} & \textbf{0.477}\sd{0.004} & \textbf{0.409}\sd{0.003} & 0.452\sd{0.015} & 0.488\sd{0.009} & \textbf{0.399}\sd{0.008} & \textbf{0.456}\sd{0.006} \\
\bottomrule
\end{tblr}}
\end{table}

The local level is supervised by HGCL on per-token representations, with the writable memory propagating context to support predictions extending beyond the current window. We probe these representations through forecasting following the evaluation protocol of \citet{lee2024soft}: a frozen encoder produces per-timestep features and a per-horizon ridge regressor predicts future values. \Cref{tab:forecasting} reports MSE and MAE on ETTh1 and Electricity at horizons $\{24, 48, 168, 336, 720\}$ in univariate and multivariate modes, with SoftCLT re-trained under the windowed protocol \METHODNAME requires for in-batch negative mining (\Cref{app:forecasting}). SoftCLT (on the TS2Vec backbone) and \METHODNAME both rely solely on contrastive objectives.

\METHODNAME achieves lower MSE and MAE on three of the four (dataset, mode) averages and on 31 of the 40 per-horizon comparisons, with the largest margin on univariate ETTh1. It wins every horizon on Electricity, while SoftCLT remains stronger at short horizons on ETTh1 univariate and at long horizons on ETTh1 multivariate.

\subsection{Architecture Comparison} \label{sec:exp:arch}

\begin{table}[!t]
\centering
\footnotesize
\caption{Comparison with \METHODNAME without PCL, xLSTM and Transformer-XL (averaged across 1\% and 5\% labeled data, ten seeds; subscripts give the sample standard deviation).}
\label{tab:xlstm-comparison}
\begin{tblr}{
  colspec={l*{8}{r}},
  colsep=2.5pt,
  row{10}={bg=gray!11},
}
    \toprule
    & \SetCell[c=2]{c}{PMT} && \SetCell[c=2]{c}{PMT w/o PCL} && \SetCell[c=2]{c}{xLSTM} && \SetCell[c=2]{c}{Transformer-XL} \\
    \cmidrule[lr]{2-3} \cmidrule[lr]{4-5} \cmidrule[lr]{6-7} \cmidrule[lr]{8-9}
    Dataset & Top-1 & mF1 & Top-1 & mF1 & Top-1 & mF1 & Top-1 & mF1 \\
    \midrule
    HAR & \textbf{92.7}\sd{0.9} & \textbf{92.7}\sd{0.9} & \underline{92.2}\sd{0.7} & \underline{92.3}\sd{0.7} & 91.7\sd{0.6} & 91.7\sd{0.6} & 91.4\sd{1.0} & 91.4\sd{1.0} \\
    Epilepsy & \textbf{96.8}\sd{0.1} & \underline{94.9}\sd{0.2} & \underline{96.8}\sd{0.1} & \textbf{95.0}\sd{0.2} & 96.7\sd{0.1} & 94.8\sd{0.3} & 96.7\sd{0.1} & \underline{94.9}\sd{0.2} \\
    Wafer & \textbf{98.4}\sd{0.4} & \textbf{95.7}\sd{1.4} & \underline{98.3}\sd{0.5} & \underline{95.5}\sd{1.6} & 97.3\sd{0.4} & 92.8\sd{1.3} & 97.0\sd{0.7} & 91.8\sd{2.6} \\
    FordA & 88.6\sd{1.1} & 88.6\sd{1.1} & \underline{88.9}\sd{1.1} & \underline{88.9}\sd{1.2} & 88.8\sd{0.7} & 88.8\sd{0.7} & \textbf{89.6}\sd{0.5} & \textbf{89.5}\sd{0.5} \\
    FordB & \textbf{76.2}\sd{0.9} & \textbf{76.1}\sd{1.1} & \underline{76.2}\sd{1.5} & 75.9\sd{1.8} & 75.7\sd{0.9} & 75.6\sd{1.0} & \underline{76.2}\sd{1.3} & \underline{76.1}\sd{1.5} \\
    POC & \textbf{69.3}\sd{2.4} & \textbf{66.2}\sd{2.8} & \underline{68.5}\sd{2.4} & 65.2\sd{3.3} & 67.6\sd{1.5} & 65.3\sd{1.6} & 67.9\sd{2.3} & \underline{65.6}\sd{2.2} \\
    ElectricDevices & \textbf{63.0}\sd{1.6} & \textbf{56.7}\sd{2.5} & \underline{62.9}\sd{1.8} & 56.2\sd{2.9} & 62.8\sd{1.9} & \underline{56.4}\sd{2.8} & 62.5\sd{2.1} & 56.3\sd{2.9} \\
    \midrule
    \textbf{Average} & \textbf{83.6}\sd{0.5} & \textbf{81.6}\sd{0.6} & \underline{83.4}\sd{0.5} & \underline{81.3}\sd{0.6} & 83.0\sd{0.5} & 80.8\sd{0.5} & 83.0\sd{0.6} & 80.8\sd{0.8} \\
    \bottomrule
\end{tblr}
\end{table}

The probes in \Cref{sec:exp:global,sec:exp:midrange,sec:exp:local} establish that the multi-scale protocol shapes useful representations, but they leave open how much of that improvement requires the writable, window-aligned memory specifically. To probe this, we replace the PMA stack with two architectural alternatives that propagate context across windows without exposing a window-aligned writable memory interface: an xLSTM stack~\citep{beck2024xlstm} (recurrent hidden state) and a Transformer-XL backbone~\citep{dai2019transformer} (read-only segment cache). Tokenizer, encoder, and SSL training are kept identical to \METHODNAME. Because neither alternative exposes a memory interface that PCL can supervise, both baselines are trained with the two protocol losses they can host (HGCL and ICL). To separate the two contributions, we also train \METHODNAME with the same two losses (\METHODNAME w/o PCL in \Cref{tab:xlstm-comparison}).

As shown in \Cref{tab:xlstm-comparison}, averaged over seven benchmarks, \METHODNAME attains the strongest Top-1 accuracy (\textbf{83.6} vs.\ 83.0 for both Transformer-XL and xLSTM) and macro-F1 (\textbf{81.6} vs.\ 80.8 for both). The matched-architecture baselines remain within $1$pp on Epilepsy, FordA, FordB, and ElectricDevices, and lose $1.0$--$1.7$pp on HAR, Wafer, and POC\@. The pattern is the one the framework predicts: alternative architectures recover part of the protocol's benefit through the losses they can host, but the mid-range level supervised by PCL is not available to them by construction, and the resulting representations carry a measurable gap from \METHODNAME's. Trained with only HGCL and ICL, \METHODNAME still leads both baselines ($83.4$ vs.\ $83.0$), and adding PCL widens the gap further.
A short-run architectural ablation on HAR further confirms these structural advantages (\cf \Cref{app:supervised:har_ablation}).

\subsection{Loss Ablation and Evaluation}
\label{sec:exp:loss}

\begin{table}[!b]
  \centering
  \footnotesize
  \caption{Ablation of the individual loss components and evaluation of their weighting coefficients, averaged over $1\%$ and $5\%$ label splits and three seeds; subscripts in (a) give the sample standard deviation over seeds, which in (b) is at most $0.48$ for Top-1 and $0.83$ for mF1. Higher is better.}
  \label{tab:ablation-main}
  \begin{minipage}[t]{0.5\linewidth}\centering
    \subcaption{Loss-component ablation: FordA, HAR, POC and Wafer.}
    \label{tab:ablation-main-har}
    {\footnotesize
\begin{tblr}{
  colspec={lccc rr},
  column{2-4}={colsep=1.5pt},
  row{3,7}={fg=gray}
}
  \toprule
  Setup & $\lambda_{\text{ICL}}$ & $\lambda_{\text{HGCL}}$ & $\lambda_{\text{PCL}}$ & Top-1 $\uparrow$ & mF1 $\uparrow$ \\
  \midrule
  All $\lambda$ = 1 & \checkmark & \checkmark & \checkmark & 87.70\sd{0.34} & 86.20\sd{0.63} \\
  \midrule
  \SetCell[c=6]{l}{\emph{Leave-one-out}} \\
  \cmidrule[lr, gray]{1-6}
  $\lambda_{\text{ICL}}=0$ & $\times$ & \checkmark & \checkmark & 85.83\sd{0.38} & 83.53\sd{0.31} \\
  $\lambda_{\text{HGCL}}=0$ & \checkmark & $\times$ & \checkmark & 86.77\sd{0.22} & 85.23\sd{0.63} \\
  $\lambda_{\text{PCL}}=0$ & \checkmark & \checkmark & $\times$ & 87.69\sd{0.38} & 86.11\sd{0.49} \\
  \midrule
  \SetCell[c=6]{l}{\emph{Single-only}} \\
  \cmidrule[lr, gray]{1-6}
  ICL only & \checkmark & $\times$ & $\times$ & 87.10\sd{0.60} & 84.66\sd{0.74} \\
  HGCL only & $\times$ & \checkmark & $\times$ & 84.68\sd{0.22} & 82.46\sd{0.68} \\
  PCL only & $\times$ & $\times$ & \checkmark & 83.32\sd{0.71} & 81.08\sd{0.97} \\
  \bottomrule
\end{tblr}
}\end{minipage}\hfill\begin{minipage}[t]{0.5\linewidth}
    \centering
    \subcaption{Evaluation of $\lambda$ coefficients on FordA, HAR, POC and Wafer.}
    \label{tab:lambda_sweeps}
    {\begin{tblr}{
  colspec={l rr rr rr},
  colsep=3pt
}
\toprule
& \SetCell[c=2]{c} {HGCL} && \SetCell[c=2]{c} {PCL} && \SetCell[c=2]{c} {ICL} & \\
\cmidrule[lr]{2-3} \cmidrule[lr]{4-5} \cmidrule[lr]{6-7}
\SetCell{c} {$\lambda$} & \SetCell{c} {Top-1 $\uparrow$} & \SetCell{c} {mF1 $\uparrow$} & \SetCell{c} {Top-1 $\uparrow$} & \SetCell{c} {mF1 $\uparrow$} & \SetCell{c} {Top-1 $\uparrow$} & \SetCell{c} {mF1 $\uparrow$} \\
\midrule
0    & 86.77 & 85.23 & 87.69 & 86.11 & 85.83 & 83.53 \\
0.01 & 87.06 & 84.95 & 87.42 & 85.76 & 86.21 & 84.42 \\
0.1  & 87.33 & 85.63 & 87.47 & 85.92 & 86.44 & 84.83 \\
0.5  & 87.55 & 85.88 & \textbf{87.78} & 86.18 & 87.29 & 85.83 \\
1    & \textbf{87.70} & \textbf{86.20} & 87.70 & \textbf{86.20} & \textbf{87.70} & \textbf{86.20} \\
\bottomrule
\end{tblr}
}\end{minipage}\end{table}

\Cref{tab:ablation-main} probes the three losses jointly~\eqref{eq:total-loss} across $1\%$ and $5\%$ label splits.
Removing any single loss from a uniform-weight configuration ($\lambda{=}1$) drops Top-1 accuracy (ICL by $1.9$pp, HGCL by $0.9$pp, PCL by $0.02$pp), and single-loss setups trail by up to $4.4$pp, indicating that all three structural levels contribute on these datasets---\cf \Cref{tab:ablation-main-har}.
Although PCL's leave-one-out drop is small here---since the classification probe reads only the sequence summary, filtering PCL's direct effect on the memory states---its substantial impact is evident when memory is probed directly in the cue-retention experiment (\Cref{sec:exp:midrange}).

Sweeping individual coefficients, \Cref{tab:lambda_sweeps}, reveals stable performance, with accuracy varying minimally ($<0.4$pp for $\lambda_\text{PCL}$, $<1$pp for $\lambda_\text{HGCL}$, $<1.9$pp for $\lambda_\text{ICL}$ (expected: the sequence summary is used for classification)).
Overall, uniform weighting ($\lambda{=}1$) lands within $0.4$pp of optimally tuned setups ($87.7$ vs.\ $88.1$ Top-1 on the same three seeds), making it a strong default (\cf \Cref{app:loss-ablation}).

\subsection{Tokenization}
\label{sec:patch-and-stride-ab}

\begin{table}[tb]
  \centering
  \footnotesize
  \caption{Evaluation of patch sizes and stride on FordA/B, and Epilepsy. Training 150 epochs, using the mean 1\% and 5\% results over 3 seeds; subscripts give the sample standard deviation.}
  \label{tab:ablation-patch-stride}
  \begin{minipage}[t]{0.5\linewidth}\centering
    \subcaption{Keeping the recipe stride constant (9 on FordA/B, 3 on Epilepsy), evaluating \textit{patch sizes} as a fraction of the seq.\ length.}
    \label{tab:ablation-patch-size}
    \resizebox{\linewidth}{!}{\footnotesize
\begin{tblr}{
  colspec={l*{6}{r}},
}
    \toprule
    & \SetCell[c=2]{c}{FordA} && \SetCell[c=2]{c}{FordB} && \SetCell[c=2]{c}{Epilepsy} & \\
    \cmidrule[lr]{2-3} \cmidrule[lr]{4-5} \cmidrule[lr]{6-7}
    Patch size & Top-1 & mF1 & Top-1 & mF1 & Top-1 & mF1 \\
    \midrule
    2.5\% & 76.1\sd{2.0} & 76.1\sd{2.0} & 68.9\sd{2.5} & 67.6\sd{3.7} & \textbf{96.8}\sd{0.2} & \textbf{94.9}\sd{0.3} \\
    5\% & 85.7\sd{1.0} & 85.7\sd{1.1} & 70.3\sd{2.7} & 69.1\sd{3.8} & \underline{96.7}\sd{0.1} & \underline{94.8}\sd{0.1} \\
    7.5\% & 88.5\sd{0.6} & 88.5\sd{0.6} & \underline{76.2}\sd{1.1} & \underline{76.2}\sd{1.1} & \underline{96.7}\sd{0.0} & \underline{94.8}\sd{0.1} \\
    10\% & \underline{89.1}\sd{0.6} & \underline{89.1}\sd{0.5} & \textbf{76.9}\sd{0.5} & \textbf{76.8}\sd{0.5} & 96.5\sd{0.1} & 94.5\sd{0.2} \\
    12.5\% & \underline{89.1}\sd{0.2} & \underline{89.1}\sd{0.2} & 75.7\sd{1.4} & 75.6\sd{1.4} & 96.4\sd{0.1} & 94.2\sd{0.1} \\
    15\% & \textbf{89.5}\sd{0.2} & \textbf{89.4}\sd{0.2} & 75.8\sd{0.5} & 75.6\sd{0.6} & 96.3\sd{0.2} & 94.2\sd{0.3} \\
    \bottomrule
\end{tblr}
}\end{minipage}\hfill\begin{minipage}[t]{0.475\linewidth}\centering
    \subcaption{With the recipe patch size held constant (62 on FordA/B, 7 on Epilepsy), evaluating \textit{stride} as a fraction of the patch size.}
    \label{tab:ablation-stride}
    \resizebox{\linewidth}{!}{\footnotesize
\begin{tblr}{
  colspec={l*{6}{r}},
}
    \toprule
    & \SetCell[c=2]{c}{FordA} && \SetCell[c=2]{c}{FordB} && \SetCell[c=2]{c}{Epilepsy} & \\
    \cmidrule[lr]{2-3} \cmidrule[lr]{4-5} \cmidrule[lr]{6-7}
    Stride & Top-1 & mF1 & Top-1 & mF1 & Top-1 & mF1 \\
    \midrule
    10\% & \textbf{88.3}\sd{0.6} & \textbf{88.3}\sd{0.6} & \textbf{76.5}\sd{0.3} & \textbf{76.4}\sd{0.4} & \textbf{96.9}\sd{0.0} & \textbf{95.1}\sd{0.0} \\
    25\% & \underline{86.1}\sd{1.0} & \underline{86.1}\sd{1.0} & \underline{73.3}\sd{1.9} & \underline{72.9}\sd{2.3} & \textbf{96.9}\sd{0.1} & \textbf{95.1}\sd{0.2} \\
    50\% & 72.0\sd{1.7} & 71.8\sd{1.7} & 69.3\sd{1.1} & 69.0\sd{0.9} & \underline{96.7}\sd{0.2} & \underline{94.8}\sd{0.3} \\
    75\% & 67.9\sd{2.0} & 67.6\sd{2.0} & 63.4\sd{1.5} & 62.4\sd{1.8} & 96.6\sd{0.2} & 94.7\sd{0.3} \\
    100\% & 65.5\sd{2.3} & 64.2\sd{2.3} & 59.5\sd{1.1} & 58.6\sd{1.3} & 96.5\sd{0.1} & 94.5\sd{0.1} \\
    \bottomrule
\end{tblr}
}\end{minipage}\end{table}

Patch length and stride control how the raw waveform is presented to \METHODNAME, and we find them to be the most consequential per-dataset choices. We sweep each independently on FordA, FordB, and Epilepsy, training each configuration for $150$ epochs and reporting the mean over $1\%$ and $5\%$ label splits across $3$ seeds.
As seen in \Cref{tab:ablation-patch-size} patch length is well-behaved: across all three datasets, accuracy varies by at most $1.2$pp over a $2\times$ range of patch sizes ($7.5$--$15\%$), with a wide plateau once patches are long enough to cover the relevant motif.
\Cref{tab:ablation-stride} shows stride matters more: moving from heavily overlapping ($10\%$ of patch length) to non-overlapping ($100\%$) patches costs $23$pp Top-1 on FordA, $17$pp on FordB, and $0.4$pp on Epilepsy. Small strides are a safe default; patch length is selected within the plateau per dataset. Because both are set relative to the sequence and patch length, one setting transfers across datasets at little cost: a $12.5\%$ patch is within $0.4$--$1.2$pp of each dataset's best, and the smallest stride is best or tied on all three. A window-size ablation in \Cref{app:window_ablation} shows similarly flat behavior for $W \in [0.2K, 0.5K]$, with degradation at smaller windows and as $W \to K$.
We discuss limitations regarding tokenizer sensitivity and batch-size requirements in \Cref{app:limitations}.

\FloatBarrier
\section{Related Work}\label{sec:related_work}

\paragraph{Self-supervised learning for time-series.}
Early contrastive methods such as CPC~\citep{oord2018representation} and SimCLR~\citep{chen2020simple} learn instance-level representations: CPC predicts future latents along the timeline with a probabilistic contrastive loss, while SimCLR treats the entire sequence as a single view and ignores within-series locality.
Later work injects explicit temporal structure.
TS-TCC~\citep{ijcai2021-324} couples cross-view future prediction with contextual instance discrimination, and CA-TCC~\citep{Eldele_2023} extends it with a class-aware term for semi-supervised settings.
TNC~\citep{tonekaboni2021unsupervised} contrasts points inside vs. outside a temporal neighborhood; SoftCLT~\citep{lee2024soft} relaxes this with soft positive assignments weighted by timestamp distance and a soft-DTW similarity over instances, so phase-shifted segments contribute graded positive signal.
TS2Vec~\citep{yue2022ts2vecuniversalrepresentationtime} enforces contrastive agreement at multiple temporal resolutions, but the resolutions are produced by repeated max-pooling, so each scale is supervised on a compressed token stream.
Hierarchical self-supervision has also been pursued in the masked-modeling family: HiMTM~\citep{zhao2024himtm} uses multi-scale masked reconstruction with self-distillation for forecasting, supervising scales through reconstruction targets rather than contrastive objectives.

These methods either supervise representation at a single scale or shape it indirectly---through hierarchical pooling that aggregates intermediate detail away, or through reconstruction targets in the masked-modeling family.
Our framework supervises three scales contrastively with the token stream preserved through the backbone: HGCL at the token level, PCL at the memory level, and ICL at the sequence level.
Unlike SoftCLT's soft assignments, HGCL derives its Gaussian weights from window geometry alone, without data-space distance or DTW alignment, and keeps the negative pool disjoint from the weighted nearby tokens.
The intermediate scale is shaped by a contrastive signal applied to first-class memory representations, rather than to surrogates produced by pooling or masked reconstruction.

\paragraph{Memory-augmented transformers.}
Vanilla transformers are stateless: once a window is processed, the past must be re-read.
Transformer-XL~\citep{dai2019transformer} extends context by caching the previous segment's hidden activations and concatenating them to the next, providing a longer horizontal context that is strictly read-only.
Compressive Transformer~\citep{rae2020compressive} extends this lineage by additionally compressing older cached activations into a coarser memory, retaining the read-only character.
Set Transformer~\citep{lee2019set} and Perceiver~\citep{jaegle2021perceiver} introduce learnable latent tokens that travel vertically through layers as a fixed-size bottleneck and reset for every sequence, never carrying sample-specific state across windows.
Sequence Complementor~\citep{chen2025sequencecomplementor} appends a small number of learnable complementary sequences to a patchified encoder; these tokens are shared across samples and discarded after the encoder, not maintaining sequence-specific state.
Titans~\citep{behrouz2025titans} introduces writable persistent slots that survive across segments, but exposes them as a single global bank decoupled from any sliding-window alignment, so the memory cannot be addressed at any specific local temporal scale.

None of these architectures provide a state interface that is simultaneously sample-specific, window-aligned, writable, and exposed at a temporally meaningful scale that a learning signal can address directly.
PMT fills this gap: each window carries a writable, sample-specific memory state aligned to a sliding window, supervised directly by PCL as a first-class representation.

\paragraph{Hierarchical and long-context transformers for time-series.}
A separate line of work tackles long-context modeling for time-series through architectural choices on the encoder rather than through memory.
PatchTST~\citep{Yuqietal-2023-PatchTST} discards memory entirely, using patch tokens and vanilla self-attention so distant dependencies are supplied through an increasingly long look-back window.
Pyraformer~\citep{liu2022pyraformer} introduces pyramidal attention with multi-scale temporal nodes connected by inter- and intra-scale edges, enabling long-range modeling at low complexity.
Rough Transformers~\citep{morenopino2024rough} compress sequences via fixed signature-based tokenization, providing continuous-time inductive bias with a lightweight token sequence.

These architectures expose temporal structure at multiple scales but do so through fixed compression or pyramidal aggregation that produces read-only intermediate nodes, not representations that a contrastive learning signal can supervise independently.
PMT instead exposes the mid-range scale as a writable memory state that is supervised contrastively at the same scale, complementing rather than replacing the token and sequence-level outputs already provided by conventional transformers.

\FloatBarrier
\section{Conclusion}

We introduced a multi-scale contrastive framework supervising time-series representations at three distinct scales: token, mid-range, and sequence-level.
To realize this, we proposed the \textbf{\METHODFULLNAME (\METHODNAME)}, augmenting conventional transformers with a sample-specific, window-aligned memory that exposes the mid-range scale as a first-class representation.
\METHODNAME probes well at every targeted level: it achieves state-of-the-art low-label classification (global), the cleanest cue retention among state-propagating architectures (mid-range), and lower forecasting error than a matched contrastive baseline on three of four settings (local).
Finally, ablations isolate the unique architectural contribution of the writable mid-range interface and support the complementarity of the three objectives, with uniform loss weighting as a stable default.

\section*{Acknowledgments and Disclosure of Funding}
This work was funded by NORSAR as part of the first author's PhD project.
This work was also funded by the Research Council of Norway through Visual Intelligence, Centre for Research-based Innovation (309439), and by the AURoRA project (359216).
The computations were performed on resources provided by Sigma2 (NN8104K and NS8104K)---the National Infrastructure for High-Performance Computing and Data Storage in Norway.
We acknowledge Sigma2 for access to the LUMI supercomputer, owned by the EuroHPC Joint Undertaking, hosted by CSC (Finland) and the LUMI consortium through Sigma2, Norway.
The authors declare no competing interests.

\FloatBarrier

\clearpage
\appendix

\crefalias{section}{appendix}
\crefalias{subsection}{appendix}

\counterwithin{figure}{section}
\counterwithin{table}{section}
\counterwithin{equation}{section}
\counterwithin{algocf}{section}

\FloatBarrier
\section{\METHODNAME Details}
\label{app:pmt_details}

\subsection{PMA Details} \label{app:pma-impl}

This appendix gives the operational details of the PMA block summarized in \Cref{sec:pma}: the gating functions, attention masking, overlap aggregation, and receptive-field growth.

\begin{figure}[tb]
\centering
\resizebox{.75\linewidth}{!}{\includegraphics{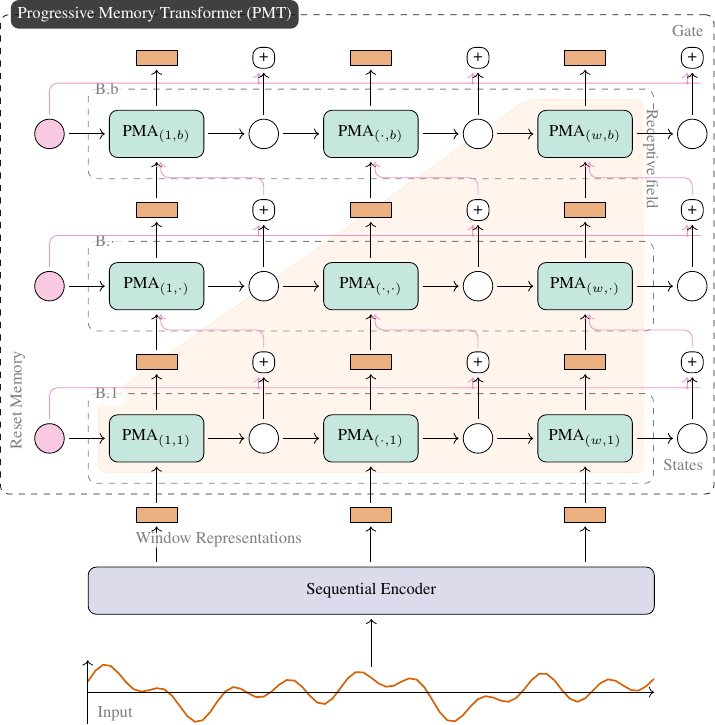}}
\caption{An illustration of \METHODNAME unrolling several PMA blocks through time (left to right) and hierarchy levels (bottom to top).
  The sequence is tokenized and encoded through an encoder.
  Then, each PMA block uses these representations and previous states to produce a new version of both of these inputs.
  At each level, a given PMA block uses a gating mechanism to control how much of the previous level's memory state is passed forward or reset (\ie, uses a reset memory state instead).}
\label{fig:pmt}
\end{figure}

The memory state propagates \emph{horizontally} to the next window and \emph{vertically} up the stack, so the receptive field widens by the stride~$S$ as the window advances while deeper blocks fuse these compact summaries without re-encoding all tokens, see \Cref{fig:pmt,fig:pma,eq:pma}.

For a given window~$w$ at a block~$b$, we compute window~$W_{w,b}$ and memory state~$M_{w,b}$ representations through our proposed PMA block, such that
\begin{equation}
    \label{eq:app-pma}
    {W}_{w,b}, {M}_{w,b} = \text{PMA} (M_{w-1, b}, \bar{M}_{w,b-1}, \bar{W}_{w,b-1}),
\end{equation}
where $M_{w-1, b}$ provides the \emph{temporal} context (memory state from the previous window at the same block), and $\bar{M}_{w,b-1}$ and $\bar{W}_{w,b-1}$ provide \emph{hierarchical} context (refined memory state and window tokens from the previous block).
The initial memory state for block~$b$ is a \emph{reset state} used for initialization and adaptive reset, \ie, $M_{0,b} = M_r$.
We refine the memory state at a given window and block with a learnable gate that mixes the carry-over memory state and the reset state (implemented as a pre-update mix that forms $\bar{M}_{w,b-1}$)
\begin{equation}
    \bar{M}_{w,b-1} = G_M(M_{w,b-1}, M_r),
\end{equation}
that adaptively mixes the memory state with the reset state~$M_r$, enabling the model to overwrite stale context and prevent drift when regimes change.
This is reminiscent of forget/reset gates in LSTMs~\citep{hochreiter1997long}, but applied to window-level memory slots updated via attention rather than to per-time-step hidden states.
Similarly, we refine the representations with an adaptive gating function,
\begin{equation}
    \bar{W}_{w,b-1} = G_W(W_{w,b-1}, W_{w,0}),
\end{equation}
that mixes the original signal representation, $W_{w,0}=f_\theta(x;w)$, with the processed one~$W_{w,b}$ to preserve local detail while integrating context.
We illustrate this block in \Cref{fig:pma}.
Within each PMA block we apply one multi-head attention layer to the concatenated input sequence~\eqref{eq:pma}.
Memory slot queries are unmasked; window queries see every key in $M_{w-1,b}$ and the strictly-causal slice of their own window but are blocked from $\bar{M}_{w,b-1}$ preventing leakage from deeper (potentially future-aware) summaries and keeping updates strictly autoregressive.
We perform this computation~\eqref{eq:pma} for every window and block in a forward manner.

In our implementation, we normalize the representations layer-wise.
Moreover, we enforce these local and causal constraints by applying custom block-diagonal attention masks.

\paragraph{Attentive overlap aggregation \& residual path.} Because windows overlap, the same position appears $O$ times per block; an overlap aggregator merges these rows so the token length stays constant across blocks.
The aggregator is a single-query cross-attention per position over its $O$ overlapped patches (keys/values from the overlaps) with a learned head$\times$overlap bias.
Its output is scaled by $1/\sqrt{O}$, post-normalized, and multiplied by a learnable $\gamma$, then fused with the masked-mean skip via a SkipGate before re-entering the residual stream.
Standard pre-norm skips wrap the chunk-processing and feed-forward sub-layers.
Additional gates control state reuse across blocks.

\subsection{Attentive Overlap Aggregator}
\label{app:overlap-aggregator}

For global position $t$ with overlapped embeddings $\{W^{(r)}_{t}\}_{r=1}^{O_t}$, the aggregator uses a \emph{single query per position} to attend over overlaps and produce
\begin{align}
A_t &= \gamma\;\mathrm{LN}\!\Bigl(O^{-1/2}\;\mathrm{Concat}_{h=1}^H\sum_{r=1}^{O_t}\alpha_{h,r}\,V_{h,r}\Bigr),\\
\alpha_{h,r}&=\mathrm{softmax}_r\!\Bigl(\tfrac{\langle q_h,K_{h,r}\rangle}{\sqrt{d_h}}+b_{h,r}\Bigr).
\end{align}
Here $K_{h,r},V_{h,r}$ are per-head projections of the overlapped $W^{(r)}_{t}$, $q_h$ is formed by depthwise Conv1D across overlaps, mean, then a 2-layer MLP, $b_{h,r}$ is a learned head\,$\times$\,overlap bias, and there is \emph{no} output projection after head concatenation. The skip path is the masked mean
\begin{equation}
S_t=\tfrac{1}{O_t}\sum_{r=1}^{O_t} W^{(r)}_{t},
\end{equation}
and the block output mixes them via a learnable SkipGate
\begin{equation}
\widehat{W}_t = G_{\text{skip}}(S_t,\,A_t).
\end{equation}
(Streaming computes the same weighted sum without materializing $[\mathcal{B},K,O,D]$ and uses a tile-constant $O$ for the $O^{-1/2}$ scaling.)

\subsection{Receptive-field growth in Progressive Memory Attention}
\label{app:rfield}

\paragraph{Notation.}
Let $W$ be the \emph{window length} (tokens per window),
$S$ the \emph{stride} ($1 \le S \le W$),
$B$ the number of stacked PMA blocks,
$K$ the total number of input tokens,
and $w\in\{1,2,\dots,N\}$ the window index,
which covers tokens $x^{\text{token}}_{(w-1)S : (w-1)S+ W-1}$.
Define
\[
\mathcal R^{(b)}_{w}\;=\;
\text{all input tokens that can influence \emph{any} token in window }w
\text{ after block }b ,
\]
so that $|\mathcal R^{(b)}_{w}| \le K$ by construction.
The bounds below report the \emph{nominal receptive-field span} induced by horizontal memory and overlap-based propagation; the actual cardinality is upper-bounded by these expressions and capped by $K$.
These bounds describe the forward history available under the PMA masking used in our experiments; if non-causal overlap aggregation is enabled, the spans should be interpreted as one-sided history bounds rather than full bidirectional receptive fields.

\begin{claim}[base case: a window in isolation]\label{claim:base}
A single window processed in isolation---without horizontal memory propagated from earlier windows---has receptive field
\begin{equation}
\label{eq:base}
\bigl|\mathcal R^{(1)}_{w}\bigr|_{\text{isolated}} = W,
\quad
\mathcal R^{(1)}_{w}\big|_{\text{isolated}}= \bigl[(w-1)S, (w-1)S+W-1\bigr].
\end{equation}
This is the receptive field consumed by the window encoder before any cross-window propagation. The two regimes below extend this base case differently.
\end{claim}

\begin{claim}[without horizontal memory]\label{claim:stateless}
If the horizontal memory bank is \emph{disabled}, only overlap-based propagation across blocks expands the receptive field; each extra block contributes at most $W-S$ additional tokens, capped by available preceding tokens,
\begin{equation}
\label{eq:no-mem}
\boxed{\;
\bigl|\mathcal R^{(b)}_{w}\bigr|_{\text{no mem}}
      \;\le\; \min\!\bigl\{K,\; W + \min\bigl((b-1)(W-S),\,(w-1)S\bigr)\bigr\}\;}
\end{equation}
which is consistent with \Cref{claim:base} at $b=1$.
\end{claim}

\begin{claim}[with horizontal memory]\label{claim:stateful}
With the memory bank active, the horizontal carry $M_{w-1,b}$ allows information from preceding windows to influence window $w$ already in the first block; subsequent blocks contribute at most $W-S$ additional tokens of nominal span via overlap propagation,
\begin{equation}
\label{eq:hor-mem}
\boxed{\;
\bigl|\mathcal R^{(b)}_{w}\bigr|_{\text{mem}}
      \;\le\; \min\!\bigl\{K,\; W + (w-1)S + (b-1)(W-S)\bigr\}\;}
\end{equation}
At $b=1$ the bound becomes $\min\{K,\, W + (w-1)S\}$, reflecting the predecessor coverage carried in via memory; \Cref{claim:base} is recovered only at $w=1$.
\end{claim}

\paragraph{FordA-sized illustration.}
Consider a token-sequence of length $K = 500$ with $W = 100$, $S = 25$, and $B = 4$. The number of valid (non-padded) windows is $N = \lfloor (K-W)/S \rfloor + 1 = 17$. For the last window ($w = 17$), the nominal span at block~1 from \eqref{eq:hor-mem} is already
\[
W + (w-1)S \;=\; 100 + 16 \times 25 \;=\; 500,
\]
so the final memory state can in principle be influenced by the whole sequence through horizontal propagation alone. With $B = 4$, the nominal span becomes
\[
W + (w-1)S + (B-1)(W-S) \;=\; 100 + 16 \times 25 + 3 \times 75 \;=\; 725,
\]
which is capped by $K$:
\[
|\mathcal R^{(4)}_{17}|_{\text{mem}} \;\le\; \min\{500,\,725\} \;=\; 500.
\]
For the EOS window the additional blocks therefore refine and recompose full-sequence information rather than enlarging the receptive set; depth contributes new uncovered tokens only for windows where the block-1 span has not yet saturated.

\subsection{Details on PCL}
\label{app:pcl-impl}

PCL is applied to the memory slots produced by the final PMA block only. Let $\mathcal{B}$ be the batch size, $v \in \{1, 2\}$ the view index, and $w = 1, \dots, N$ the window index.
The memory state of the $i$-th sample at window $w$ under view $v$ is $M_{i, w}^{(v)} \in \mathbb{R}^{n_m \times D}$, with $n_m$ slots per window.
For an anchor $m_a = M_{i, w}^{(1)}[k]$, the positive is $m_p = M_{i, w}^{(2)}[k]$, and the loss averages over all anchor slots with normalization $1/(\mathcal{B} N n_m)$.
The negative pool $\mathcal{N}_a$ excludes all slots from the same source sequence as the anchor; we typically sample $n_\text{neg} = 512$ negatives per anchor.

Symmetry is enforced by computing the loss in both anchor/key directions and averaging: $\mathcal{L}_\text{PCL} = \tfrac{1}{2}(\mathcal{L}_\text{PCL}^{1 \to 2} + \mathcal{L}_\text{PCL}^{2 \to 1})$. The slots are $\ell_2$-normalized and used as contrastive embeddings. The temperature $\tau$ is dataset-dependent and reported with the per-dataset hyperparameters in our released code. The loss is computed in chunks over anchors to bound activation memory, with a custom autograd path that stores compact backward statistics rather than retaining all sampled negatives. The composition of the negative pool, the number of negatives and an optional per-slot projection head are likewise set per dataset.

\subsection{Details on HGCL}
\label{app:hgcl-impl}

HGCL applies a bucket-form InfoNCE objective at two scales---token-within-window and window-across-sequence.
Both scales operate on the same projected representations $z^{(v)} \in \mathbb{R}^{\mathcal{B} \times K \times D}$ (view $v \in \{1, 2\}$, $K$ tokens after patchification), unfolded into windows of length $W$ and stride $S$.

\paragraph{Bucket objective.}
For an anchor $a$ with aligned positive $p$, kernel-weighted aggregate $T_a = \sum_{c} q_{a,c}\, s_{a,c}$ over local candidates $c$, and negative bank $\mathcal{N}_a$
\begin{equation}
    \label{eq:hgcl-bucket}
    \mathcal{L}_a = - \log \frac{\exp(s_{a,p} / \tau)}{\exp(s_{a,p} / \tau) + \exp((T_a - s_{a,p}) / \tau) + \sum_{n \in \mathcal{N}_a} \exp(s_{a,n} / \tau)},
\end{equation}
where $s_{a,c} = \langle z_a, z_c \rangle$ is cosine similarity and $\tau$ is a level-specific temperature.

\paragraph{Conventions.} The kernel weights $q_{a,c}$ are unnormalized Gaussian values; $q_{a,p}=1$ at the aligned candidate, and $T_a$ sums over all local candidates including $p$, so $T_a - s_{a,p} = \sum_{c\neq p} q_{a,c}\, s_{a,c}$. Subtracting $s_{a,p}$ keeps the aligned positive in the numerator without double-counting in the denominator.

\paragraph{Token level.}
For an anchor token at position $k$ in window $w$ of view~$1$, the aligned positive is the matched token at the same position in view~$2$. Local candidates are the $W$ tokens of the same window in view~$2$, with Gaussian weights
\begin{equation}
    q_{k, j}^{t} = \exp\!\left(-\frac{(k - j)^2}{2 \sigma_p^2}\right), \qquad \sigma_p \propto W.
\end{equation}
Negatives are sampled from tokens of other sequences in the global batch.

\paragraph{Window level.}
Window representations are the $\ell_2$-normalized mean of their tokens. For an anchor window at position $u$ in view~$1$, the aligned positive is the matched window in view~$2$. Local candidates are the windows of the same sequence in view~$2$, with Gaussian weights
\begin{equation}
    q_{u, u'}^{w} = \exp\!\left(-\frac{((u - u') \cdot S)^2}{2 \sigma_w^2}\right), \qquad \sigma_w \propto W \cdot \frac{W - S}{W}.
\end{equation}
Negatives are sampled from windows of other sequences in the global batch.

\paragraph{Combined loss.}
The total HGCL loss combines the two scales
\begin{equation}
    \mathcal{L}_\text{HGCL} = \alpha\, \overline{\mathcal{L}_t} + \beta\, \overline{\mathcal{L}_w},
\end{equation}
where $\overline{\,\cdot\,}$ denotes mean over valid anchors.

\subsection{Details on ICL}
\label{app:icl-impl}

The loss in Eq.~\eqref{eq:icl} is computed symmetrically over view assignments
\begin{equation}
    \mathcal{L}_\text{ICL} = \tfrac{1}{2}\bigl(\mathcal{L}_\text{ICL}^{1 \to 2} + \mathcal{L}_\text{ICL}^{2 \to 1}\bigr).
\end{equation}
The projection head $g(\cdot)$ is a two-layer MLP with GELU activation, applied per view.

\FloatBarrier
\section{Implementation Details}
\label{app:implementation}

\begin{figure}[tb]
    \LinesNotNumbered
    \footnotesize
\scriptsize
\begin{algorithm}[H]
  \caption{\textbf{Forward pass through a $B$-block PMA stack}}
  \label{alg:pma-stack}

  \textbf{Input:} patch tokens $W_{1:N,0}$, random memory $M_r$ \\
  \textbf{Output:} token stream $W_{1:N,B}$, memories $M_{1:N,B}$

  \BlankLine
  \For{$b\gets1$ \KwTo $B$}{$M_{0,b}\!\leftarrow\!M_r$\;
    \For{$w\gets1$ \KwTo $N$}{$\bar{W}_{w,b-1}\!\leftarrow\!
         G_W\bigl(W_{w,b-1}, W_{w,0}\bigr)$;
      $\bar{M}_{w,b-1}\!\leftarrow\!
         G_M\bigl(\mathrm{LN}(M_{w,b-1}),\mathrm{LN}(M_r)\bigr)$;
      $\bigl[M_{w,b},W^{\mathrm{out}}_{w,b}\bigr]\!\leftarrow\!
         \mathrm{SelfAttn}\!\bigl([M_{w-1,b}\|\bar{M}_{w,b-1}\|\bar{W}_{w,b-1}]\bigr)$;}
    $W_{w,b}\!\leftarrow\!
       \mathrm{OverlapPool}\bigl(\{W^{\mathrm{out}}_{w',b}\}_{w'=1}^N\bigr)\;(\forall w)$}
\end{algorithm}

\vspace{-3pt}
\begin{itemize}[leftmargin=1.2em,itemsep=0pt]
  \scriptsize
  \item $G_W$ --- mixes new patch evidence with prior tokens.
  \item $G_M$ --- gate blending inherited memory with $M_r$.
  \item $\mathrm{SelfAttn}$ --- masked attention over $[M_{w-1,b}\|\bar{M}_{w,b-1}\|\bar{W}_{w,b-1}]$.
  \item $\mathrm{OverlapPool}$ --- attentive overlap aggregator producing $W_{w,b}$.
\end{itemize}

    \caption{Forward pass through a stack of $B$ PMA blocks: per-window, per-block memory and token updates with adaptive gates and asymmetric attention masking.}
    \label{app:pma_forward}
\end{figure}

We train the PMT using an Adam optimizer with a cosine annealing learning rate~\citep{loshchilov2017sgdr} scheduler; learning rates, warmup, and the gates' initialization and learning-rate multiplier are set per dataset in the released configurations.
Models were trained with a batch size of 256.
The models were trained on Nvidia GH200 GPUs.
For dataset specific hyperparameters used for Table~\ref{tab:classification_table}, we refer the reader to the attached repository (link on page 1).

The cosine similarity figures were all created using per-dataset specific frozen backbones, whose configurations are included in the released code.
For the HAR figures, the backbone consisted of 6 PMA blocks with a window size of 6 and stride 3, using 2 memory states; the Epilepsy backbone uses a window size of 12 and stride 6.
Commonly for all visualizations, we do a simple forward pass through the model and extract the tokenizer tokens that enter the PMA blocks and the final PMA block's memory states.
We use the mean per window memory state for the visualizations.
Both the averaged memory state and the tokens are $\ell_2$-normalized, after subtracting the mean representation at each position over the validation set (likewise for the TS2Vec embeddings in \Cref{fig:softclt_fig4}).

The scatter plots use t-SNE (\Cref{fig:pca-repr-standing}) or PCA (\Cref{fig:pca-repr}) for dimensionality reduction per token and per memory state.
Heat-maps, such as \Cref{fig:hybrid-sims,fig:sim-dissim} use the cosine-similarity matrix for both the token and memory state visualizations.

\paragraph{Patchified input.} A 1-D convolution first tokenizes the waveform into fixed-length patch embeddings; these tokens---not the raw samples---form the input to every PMA block.
Although we restrict ourselves to this patchified view in the present work, extending PMA to operate directly on raw time steps is a promising avenue for future research.

\paragraph{Overlap-aware processing.} We unfold the patch stream into windows of length $W$ and stride $S$ (overlap $O=\lceil W/S\rceil$).
Each window passes through a single-layer Transformer with the asymmetric mask described in \Cref{app:pma-impl}; a fused attention kernel reduces its space requirement to $\mathcal{O}\bigl((|M|+S)D\bigr)$ per window.
After all $N$ windows are processed, the attentive overlap aggregator merges the $O$ overlapping rows at every position.
Without this overlap aggregator each subsequent PMA block would receive a growing number of tokens due to $W > S$.
The overlap aggregator ensures the input and output shape of any PMA block remains equal.

\paragraph{Global aggregation.} Finally, we append the \cls token to the output from the PMA blocks, and pass the tokens through a series of encoder layers with neighborhood masking.
Specifically, a patch token at position $t$ attends only to its local context window (positions $t-n, \dots, t$), whereas the \cls token attends to every position to aggregate the global representation.

\FloatBarrier
\section{Computational Analysis}\label{app:compute}
We report minimal, reproducible compute measurements for completeness.
These results detail compute and resource notes already in \Cref{app:implementation} and use the same backbone as the experiments.
The goal is to characterize the overhead of PMA relative to a vanilla transformer with FlashAttention in the \textit{same} implementation, rather than to claim PMT is more efficient than long-context architectures such as Transformer-XL~\citep{dai2019transformer} or patch-based forecasters like PatchTST~\citep{Yuqietal-2023-PatchTST}.
We distinguish between a \emph{streaming} PMA microbenchmark (single-window kernel; no re-encoding the past) and a \emph{vectorized} non-streaming configuration (overlapped windows materialized for speed during training).

\subsection{Hardware and Setup}\label{app:resources}

The main paper's experiments were trained on NVIDIA GH200 ($\sim$95\,GiB) GPUs---typically 2 GPUs via DDP, with 8 dataloader workers per GPU.
A full $250$-epoch experiment under this configuration takes roughly 1.5 hours.
The streaming and vectorized benchmarks reported in \Cref{app:compute:stream,app:compute:vectorized,app:compute:aggregator} were run separately on a single NVIDIA GH200 in PyTorch, in FP16 (FP32 for the vectorized configuration); they characterize PMA's cost relative to full attention in the same implementation rather than absolute throughput.
Peak GPU memory is measured via \texttt{torch.cuda.max\_memory\_allocated()} after \texttt{cudaDeviceSynchronize()}; on shorter sequences, different models report nearly the same peak because a common floor of weights, inputs and workspaces dominates, and we report the measured peak in all cases.

\subsection{Streaming PMA microbenchmarks}\label{app:compute:stream}
We time a single PMA block with $D=320$ (as in the main experiments), a 16-sample convolutional patchifier (8$\times$ compression), a window of $20\%$ of the token sequence, window stride $S=0.5\,W$, and one memory slot.
For comparability, full attention is run on the full sequence (global receptive field), both as dense attention and with FlashAttention (FA).
We report per-window latency (ms), total sequence latency (ms) implied by sliding over the sequence, maximum sustained sampling rate (Hz) with a 20\% head-room, and peak allocated memory.

\begin{table}[tb]
  \centering
  \footnotesize
  \caption{Streaming PMA: single-window kernel timed; full attention run on the full sequence, dense and with FlashAttention (FA). PMA's per-window cost stays essentially flat over this range (a fixed overhead dominates), and its peak memory stays within that of FA where dense attention grows quadratically; 96\,kHz real-time with a $20\%$ head-room is sustained from 4,096 samples upward.}
  \label{tab:streaming-pma}
  \resizebox{\linewidth}{!}{\begin{tblr}{
  colspec={l*{7}{r}},
}
\toprule
\SetCell[r=2]{l,valign=f} Sequence (samples $\to$ tokens) & \SetCell[r=2]{r,valign=f} PMA window ms (seq ms) & \SetCell[c=2]{c} Full sequence [ms] && \SetCell[r=2]{r,valign=f} Max SR [Hz] $\uparrow$ & \SetCell[c=3]{c} Peak memory [MiB] && \\
\cmidrule[lr]{3-4}\cmidrule[lr]{6-8}
& & Dense & FA & & PMA & Dense & FA \\
\midrule
$512 \to 63$ & 1.0 (9.0) & 0.5 & 0.5 & 22{,}050 & 36 & 36 & 36 \\
$1{,}024 \to 127$ & 1.7 (15.3) & 0.6 & 0.5 & 44{,}100 & 36 & 36 & 36 \\
$2{,}048 \to 255$ & 1.7 (15.3) & 0.5 & 0.5 & 48{,}000 & 36 & 37 & 37 \\
$4{,}096 \to 511$ & 1.6 (14.7) & 0.5 & 0.5 & 96{,}000 & 37 & 41 & 39 \\
$8{,}192 \to 1{,}023$ & 1.7 (15.2) & 0.5 & 0.5 & 96{,}000 & 38 & 55 & 43 \\
$16{,}384 \to 2{,}047$ & 1.7 (15.2) & 0.5 & 0.5 & 96{,}000 & 41 & 106 & 45 \\
$32{,}768 \to 4{,}095$ & 1.7 (15.1) & 1.2 & 0.5 & 96{,}000 & 47 & 304 & 55 \\
$65{,}536 \to 8{,}191$ & 1.6 (14.8) & 4.2 & 0.9 & 96{,}000 & 66 & 1{,}084 & 75 \\
$131{,}072 \to 16{,}383$ & 1.7 (14.9) & 17.0 & 2.6 & 96{,}000 & 115 & 4{,}181 & 115 \\
\bottomrule
\end{tblr}
}
\end{table}

\Cref{tab:streaming-pma} shows that over this range PMA's per-window latency is dominated by a fixed overhead and stays essentially flat as the window grows with the sequence; dense full attention's latency grows quadratically and exceeds PMA's implied sequence latency at the longest horizon, while FA remains faster across the whole range.
Streaming PMA keeps the per-step memory bounded by $O((|M|+S)D)$ (\Cref{app:implementation}), which avoids the quadratic growth of dense full-sequence attention; FA's memory is also linear in sequence length, and the two meet at the longest horizon.

\subsection{Vectorized non-streaming inference}\label{app:compute:vectorized}
For training speed, we also report a non-streaming vectorized configuration that materializes all overlapped windows before the attentive overlap aggregator.
All models use $D=320$, 6 blocks (PMT includes three neighborhood encoders for \cls), 100{,}000 time steps (8$\times$ tokenization), $W=0.1K$, $S=0.5W$, one memory slot.
Latency is end-to-end.

\begin{table}[tb]
    \centering
    \footnotesize
    \caption{Non-streaming inference (vectorized). Vectorization duplicates overlapped windows for speed, inflating peak memory; the attentive overlap aggregator removes duplicates post-block.}
    \label{tab:vectorized}
\begin{tblr}{
  colspec={lcccc},
}
\toprule
Model & Params [M] & Peak mem [MiB] & Latency [ms] $\downarrow$ & Tokens/s $\uparrow$ \\
\midrule
Full attention       & 7.4 & 8{,}361 & 1{,}911.81 & 209{,}209 \\
PMA (vectorized)     & 16.3 & 10{,}872 & 1{,}440.39 & 277{,}680 \\
PMT (full)           & 20.1 & 11{,}813 & 2{,}607.74 & 153{,}377 \\
\bottomrule
\end{tblr}
\end{table}

\paragraph{Notes and caveats.} (i)~FA sees the full sequence at once whereas streaming PMA strictly limits the receptive field to the current window plus memory slots; this explains FA's lower latency and PMA's memory advantage over dense attention on long horizons.
(ii)~The vectorized PMA/PMT inflate peak memory due to overlapped materialization; the streaming kernel avoids this by construction.

\subsection{Attentive Overlap Aggregator Cost}\label{app:compute:aggregator}

To understand the computational distribution within PMA blocks, we isolate and measure the \emph{attentive overlap aggregator} component at the window geometries of the streaming and vectorized benchmarks.

\paragraph{Methodology.}
We measure the aggregator in isolation by providing the pre-computed window embeddings captured from a full forward pass of the block on the same inputs, and report its share of that block's time.
Setup matches \Cref{app:compute:stream}, with one memory slot.

\begin{table}[tb]
  \centering
  \footnotesize
  \caption{Attentive overlap aggregator: share of a full PMA block's forward time, with the block and the aggregator timed on identical inputs (FP16, batch 1, one memory slot, $d{=}320$, $S{=}W/2$).}
  \label{tab:agg-streaming}
\begin{tblr}{
  colspec = {l*{5}{r}},
}
\toprule
Geometry & $W$ (tokens) & $L$ (tokens) & Block [ms] & Aggregator [ms] & Time share [\%] $\downarrow$ \\
\midrule
\SetCell[r=3]{l} $W{=}0.2L$ & 1{,}024 & 5{,}120 & 11.3 & 1.5 & 12.9 \\
 & 2{,}048 & 10{,}240 & 12.6 & 2.7 & 21.2 \\
 & 4{,}096 & 20{,}480 & 15.4 & 5.3 & 34.5 \\
\midrule
\SetCell[r=3]{l} $W{=}0.1L$ & 2{,}500 & 25{,}000 & 28.0 & 6.5 & 23.2 \\
 & 5{,}000 & 50{,}000 & 39.8 & 12.8 & 32.2 \\
 & 10{,}000 & 100{,}000 & 101.4 & 25.5 & 25.1 \\
\bottomrule
\end{tblr}
\end{table}

\paragraph{Results.}
\Cref{tab:agg-streaming} shows that the overlap aggregator accounts for 13--35\% of a PMA block's forward time across these geometries.
The aggregator's computational cost is dominated by key-value projections (87\% of aggregator FLOPs by analytic count), while the attention mechanism itself requires minimal computation due to single-token queries over $O{=}2$ positions.

\paragraph{Implementation notes.}
Our analysis isolates the aggregator component to measure its inherent computational cost.
The vectorized implementation materializes overlapped windows for training speed but increases peak memory.

\subsection{\texorpdfstring{Per-architecture training and inference cost}{Per-architecture training and inference cost}}\label{app:compute:cost}
\Cref{tab:arch-cost} compares the training and inference cost of \METHODNAME with the matched xLSTM and Transformer-XL baselines of \Cref{sec:exp:arch} on one NVIDIA GH200, at the recipes of \Cref{tab:xlstm-comparison}. \METHODNAME trains as fast as or faster than xLSTM on five of the seven datasets and uses less peak memory on all seven; Transformer-XL is the lightest of the three.
\begin{table}[tb]
  \centering
  \footnotesize
  \caption{Training and inference cost on one NVIDIA GH200. \METHODNAME and the matched xLSTM and Transformer-XL baselines share each dataset's tokenizer, input stem and width and train with HGCL and ICL; \METHODNAME adds PCL. Train step: median time of a full BF16 optimizer step at batch 256 on synthetic inputs; peak memory: maximum allocated CUDA memory during those steps; inference: FP32 sequences per second at batch 256. xLSTM's sLSTM layers run on the library's PyTorch backend.}
  \label{tab:arch-cost}
\begin{tabular}{llrrrr}
\toprule
Dataset & Architecture & Params (M) & Train step (ms) & Peak memory (GB) & Inference (seq/s) \\
\midrule
HAR & PMT & 28.1 & 513 & 25.2 & 988 \\
 & xLSTM & 10.2 & 1{,}428 & 98.5 & 605 \\
 & Transformer-XL & 18.4 & 206 & 18.5 & 7{,}265 \\
\addlinespace
Epilepsy & PMT & 18.7 & 573 & 11.3 & 2{,}335 \\
 & xLSTM & 8.0 & 784 & 45.2 & 1{,}239 \\
 & Transformer-XL & 12.2 & 69 & 5.0 & 27{,}458 \\
\addlinespace
Wafer & PMT & 18.0 & 662 & 10.1 & 1{,}906 \\
 & xLSTM & 7.4 & 673 & 38.4 & 1{,}530 \\
 & Transformer-XL & 11.5 & 67 & 4.2 & 28{,}184 \\
\addlinespace
FordA & PMT & 21.8 & 213 & 11.2 & 4{,}317 \\
 & xLSTM & 14.7 & 528 & 31.4 & 1{,}916 \\
 & Transformer-XL & 17.5 & 81 & 6.5 & 22{,}461 \\
\addlinespace
FordB & PMT & 19.4 & 208 & 10.7 & 4{,}373 \\
 & xLSTM & 12.3 & 522 & 30.6 & 1{,}928 \\
 & Transformer-XL & 15.1 & 74 & 5.7 & 24{,}364 \\
\addlinespace
POC & PMT & 22.4 & 428 & 6.0 & 3{,}180 \\
 & xLSTM & 13.6 & 330 & 17.9 & 3{,}090 \\
 & Transformer-XL & 17.0 & 69 & 3.0 & 22{,}834 \\
\addlinespace
ElectricDevices & PMT & 24.2 & 330 & 7.3 & 2{,}672 \\
 & xLSTM & 13.5 & 258 & 13.2 & 3{,}934 \\
 & Transformer-XL & 17.7 & 64 & 2.2 & 21{,}107 \\
\bottomrule
\end{tabular}
\end{table}

\section{Experimental Protocol and Datasets}\label{app:protocol}

\subsection{Datasets}\label{app:datasets}

In our experiments, we rely on a set of well-established time-series benchmarks drawn from the UCR, UEA, and UCI repositories~\citep{UCRArchive2018, UEAArchive2018}.
The Human Activity Recognition (\textbf{HAR}) dataset~\citep{anguita2013public} contains triaxial accelerometer and gyroscope streams recorded at 50 Hz from 30 volunteers who carried a smartphone on their waist while performing six everyday actions (walking, climbing or descending stairs, sitting, standing, and lying).
For epileptic-seizure detection we adopt the version of the \textbf{Epilepsy} dataset simplified by TS-TCC: the original EEG collection---23.6-second segments from 500 subjects and five classes~\citep{andrzejak2001indications}---is reduced to a binary seizure/non-seizure task.
The remaining benchmarks---Wafer, FordA, FordB, PhalangesOutlinesCorrect (POC), ElectricDevices, and StarLightCurves---are sourced from the UCR archive~\citep{UCRArchive2018}.
\textbf{Wafer} contains inline process-control sensor traces from silicon-wafer fabrication and is strongly imbalanced: defective wafers constitute 9.7\% of the training set and 10.8\% of the test set.
\textbf{FordA} and \textbf{FordB} each comprise 500-sample engine-vibration sequences used to decide whether a specific subsystem fault is present; FordA was recorded under controlled laboratory noise, whereas FordB reflects normal operating conditions.
The \textbf{POC} dataset merges three tasks derived from more than 1,300 radiographs employed for bone-age estimation, with labels indicating whether the automatically extracted phalange outlines are correct.
\textbf{StarLightCurves} contains phase-folded light curves of variable stars from astronomical surveys in three classes; at 1,024 samples per sequence it is the longest dataset in our suite.
Finally, the \textbf{ElectricDevices} dataset contains electricity-consumption profiles from 251 UK households, gathered to study residential usage patterns and help lower carbon emissions.
Detailed statistics for all datasets appear in Table~\ref{tab:self_supervised_datasets}.

\begin{table}[tb]
  \centering
  \footnotesize
  \caption{Datasets used for self-supervised classification}
  \label{tab:self_supervised_datasets}
  \begin{tblr}{
    colspec={lrrrrr}
  }
    \toprule
    {Dataset} & {\# Train} & {\# Test} & {Length} & {\# Channel} & {\# Class} \\
    \midrule
    HAR              &  7\,352 &  2\,947 & 128  & 9 & 6 \\
    Epilepsy         &  9\,200 &  2\,300 & 178  & 1 & 2 \\
    Wafer            &  1\,000 &  6\,164 & 152  & 1 & 2 \\
    FordA            &  3\,601 &  1\,320 & 500  & 1 & 2 \\
    FordB            &  3\,636 &    810  & 500  & 1 & 2 \\
    POC              &  1\,800 &    858  &  80  & 1 & 2 \\
    ElectricDevices  &  8\,926 &  7\,711 &  96  & 1 & 7 \\
    StarLightCurves  &  1\,000 &  8\,236 & 1\,024 & 1 & 3 \\
    \bottomrule
  \end{tblr}
\end{table}

\paragraph{Selection.} The eight datasets span six domains, sequence lengths from 80 to 1,024, and channel counts from 1 to 9, providing diversity along the axes (length, channelization, class cardinality) most relevant to contrastive pretraining. Their training splits jointly comprise $\sim$15.9M channel time-steps, sufficient for in-batch InfoNCE training without external memory banks or queues~\citep{he2020momentum,wu2018unsupervised}, which lets us use a comparable training recipe across all datasets. Micro-datasets with only tens of sequences (e.g., \textit{GunPoint}, \textit{CBF}) cannot accommodate this batch-size requirement and are excluded.

\subsection{\texorpdfstring{Evaluation Protocol}{Evaluation Protocol}}\label{app:linear_eval}

For classification (\Cref{sec:exp:global}) we follow the evaluation protocol of \citet{ijcai2021-324}: the backbone is pretrained on each training set without labels using the multi-scale contrastive objective of \Cref{sec:method}, then frozen. An SVM classifier with an RBF kernel, as in \citet{lee2024soft}, is trained on 1\% and 5\% labeled subsets, and top-1 accuracy and macro-F1 are reported on the official test split. Probes are fit periodically during pretraining, and we report the one with the highest top-1 accuracy on that split, together with its macro-F1; on five of the eight datasets, the SVM's $C$ and $\gamma$ are selected on that split as well. Baseline results are those reported by \citet{lee2024soft}. Per-dataset hyperparameters are documented in the released code.

\subsection{Cue Retention Protocol} \label{app:retention}

\paragraph{Why this experiment.}
Memory-augmented and state-propagating architectures expose intermediate states that are claimed to carry information across windows, but downstream task accuracy alone leaves two factors entangled: the architecture's capacity to carry information through its state, and the training objective's role in shaping what gets stored there.
Downstream accuracy is also a poor instrument for either: many classification benchmarks are decidable from local features alone, so a model with an inert intermediate state can score competitively without ever propagating mid-range information, and a model that propagates well can underperform on tasks that don't reward it.
The cue-retention probe is designed to bypass this conflation by introducing a controlled, out-of-distribution perturbation into the input and asking only whether evidence of that perturbation survives propagation through the model's state. The probe is OOD by construction---the model is never trained on the cue---so success requires retention as an architectural and supervisory property, not as a task-specific shortcut.

\paragraph{Application to \METHODNAME.}
We use the protocol to disentangle two contributions specific to our design: the writable, window-aligned memory \emph{interface} (architecture) and the PCL objective that supervises it (training).
The architectural contribution is measured directly by the $\lambda_{\text{PCL}}=0$ row, in which \METHODNAME retains the writable memory mechanism but receives no contrastive signal at the memory level.
The supervisory contribution is recovered by the gap between $\lambda_{\text{PCL}}=0$ and full \METHODNAME. Read this way, \Cref{tab:forda-cue-eos} reports two distinct quantities for our model in the same table: how much mid-range capacity the architecture provides on its own, and how much PCL adds on top.

\paragraph{Cue construction.}
A Hann-tapered burst is added to a single input channel, with amplitude $\alpha \sigma$ where $\sigma$ is the channel's standard deviation computed \emph{excluding} the cue span (preventing leakage through normalization), and $\alpha = 5$.
The burst width is fixed at 26 samples across all runs and architectures.
The burst is placed at a window offset chosen from a configured set of gap ratios, controlling the distance from cue to end of sequence.
Cue retention is run with a non-overlapping-window variant of PMA for compute reasons; the variant shares the writable, PCL-supervised memory interface evaluated elsewhere in the paper.
The models are trained on FordA \emph{without any exposure to the cue}; the cue is introduced only at evaluation time on the test set; and the backbone remains frozen throughout. Results are averaged over five pretraining seeds, each evaluated at its final checkpoint.

\paragraph{Probe.}
We use the \textbf{end-of-sequence (EOS)} probe: features are extracted only from the final window per sequence, and a logistic-regression classifier (with standardized features) predicts whether the sequence contained a cue at all.
Restricting to a single fixed-position feature per sequence prevents the probe from exploiting positional correlations introduced by the gap-ratio sampling. Cues are sampled at gap ratios $\{0, 0.25, 0.5, 0.75\}$. At ratio $0$ the cue falls in the final window; at the others the probe can succeed only if its evidence has been retained through state propagation. Over those upstream gaps, EOS AUC is $0.994$ for \METHODNAME, $0.982$ for xLSTM, $0.960$ for \METHODNAME-$\lambda_{\text{PCL}}{=}0$, $0.956$ for Transformer-XL, and $0.514$ for TS2Vec.

\paragraph{Metrics.} Let $g \geq 0$ index the gap (in windows) between the cue and the end of sequence; sequences without a cue contribute the negative class.
\begin{itemize}[leftmargin=*, nosep, topsep=0pt]
    \item \textbf{EOS AUC.} Standard binary ROC-AUC over the EOS probe's predicted cue-presence probabilities. Threshold-free; primary headline metric.
    \item \textbf{EOS Top-1.} Top-1 accuracy of the EOS probe at the natural decision threshold ($p=0.5$).
\end{itemize}
Both are averaged across seeds; the probe and feature extractors are otherwise identical across architectures, so AUC differences reflect representation quality, and the threshold-based gap between AUC and Top-1 reflects how well the representation places the decision boundary at a usable threshold.

\paragraph{Feature extraction.}
Per architecture, the EOS feature is the final-window state representation each architecture exposes natively: for \METHODNAME, the final PMA block's memory states at the last window (mean-pooled across slots when more than one is configured); for Transformer-XL, the segment cache at the final position; for xLSTM, the recurrent hidden state at the final time step.
TS2Vec has no native window-aligned state, so we unfold its intermediate convolutional embeddings into windows matching \METHODNAME's window length and stride and mean-pool within each window. Titans~\citep{behrouz2025titans} is not included here or in \Cref{sec:exp:arch}: its memory is a single global bank rather than window-aligned (\Cref{sec:related_work}), and without a reference implementation a faithful matched comparison was not possible.

\paragraph{Supervisory signal.}
For \METHODNAME the EOS feature is the final memory state; for the matched-architecture baselines (\Cref{sec:exp:arch}) the corresponding state representation; for TS2Vec, the windowized intermediate embeddings (the closest analogue in a backbone without explicit window-aligned state).
To separate the writable-memory \emph{interface} (architecture) from the effect of supervising it directly, we also report \METHODNAME with $\lambda_{\text{PCL}}=0$.

\subsection{Forecasting Protocol}\label{app:forecasting}

\METHODNAME follows the forecasting evaluation protocol of \citet{lee2024soft}: a frozen encoder produces per-timestep representations, and a per-horizon ridge regressor predicts future values on ETTh1~\citep{zhou2021informer} and Electricity~\citep{trindade2015electricity} at horizons $\{24, 48, 168, 336, 720\}$, using the same data splits as SoftCLT.
The main difference is that \METHODNAME relies on windowed training to support in-batch negative mining, which requires preparing the forecasting data as fixed windows rather than full series with random temporal crops.
To ensure comparability, we retrain SoftCLT under this adapted windowed protocol while keeping their architecture and forecasting hyperparameters fixed; results in \Cref{tab:forecasting} use these retrained SoftCLT numbers.
Both methods train for $5{,}000$ updates (batch size $128$ for \METHODNAME and $8$ for SoftCLT). Each training view of \METHODNAME masks timesteps independently with probability $0.5$; the loss weights are $\lambda_\text{HGCL}{=}1$, $\lambda_\text{PCL}{=}0.1$, and $\lambda_\text{ICL}{=}0$ in all four settings. For each method, setting, and seed, the checkpoint with the lowest validation MAE is frozen before test, and the ridge penalty is selected on validation for each horizon from a grid of $23$ values ($10^{-3}$ to $10^{5}$); using SoftCLT's grid of $13$ values changes no winner in \Cref{tab:forecasting}.

\paragraph{Causal feature extraction.}
Per-timestep features are produced via the sliding-encode convention of \citet{yue2022ts2vecuniversalrepresentationtime, lee2024soft}: the feature at original time $t$ is the backbone output at the final position of a separate forward pass over the left-padded input window $[\max(0,\,t{-}P),\,t]$, with $P=200$. By construction, no input at position $>t$ participates in the computation of the feature at $t$, regardless of any non-causal attention internal to the backbone. For forecasting we configure the tokenizer with patch size 1 and stride 1 so that token positions align one-to-one with timesteps; this differs from the patch-based tokenizer used for classification (\Cref{sec:patch-and-stride-ab}). Left padding uses NaN values, which the backbone treats as masked positions.

\FloatBarrier
\section{Additional Experimental Results}
\label{app:additional}

\subsection{Supervised Results and Additional Ablations}
\label{app:supervised}

To verify that the architecture is not limited to SSL, we train \METHODNAME end-to-end with \emph{full supervision} (cross-entropy on the label) using the same tokenizer and encoder backbone as in the main text.
The sequence-level representation is passed to a linear classification head.
Unless noted, optimization and scheduling follow \Cref{app:implementation} (Adam + cosine decay).
For the fully supervised experiments in this section, no self-supervised losses are used, and we report the epoch with the highest top-1 accuracy, together with its macro-F1.

\subsection{Full-Supervision (Cross-Entropy) Results}
\label{app:supervised:full}
\Cref{tab:supervised_full} reports Top-1 accuracy and macro-F1 on four representative datasets.
Results show that \METHODNAME matches or exceeds a vanilla Transformer trained with the same supervised protocol on all four datasets (by up to $2.4$pp, on FordB).
These results confirm the \emph{task-agnostic} nature of the backbone: while the main paper focuses on low-label SSL, the same architecture trains effectively in a purely supervised regime.

\begin{table}[tb]
  \centering
  \footnotesize
  \caption{\textbf{Fully supervised results} (cross-entropy, 250 epochs). Top-1 / macro-F1 (\%) on the official test split; mean and standard deviation over three seeds.}
  \label{tab:supervised_full}
  \begin{tblr}{
    colspec={l*{4}{r}}
  }
    \toprule
    \SetCell[r=2]{l,valign=f} Dataset & \SetCell[c=2]{c} \METHODNAME (supervised) && \SetCell[c=2]{c} Vanilla Transformer (supervised) & \\
    \cmidrule[lr]{2-3}\cmidrule[lr]{4-5}
    & Top-1 & mF1 & Top-1 & mF1 \\
    \midrule
    HAR   & 94.6\sd{0.1} & 94.6\sd{0.1} & 94.3\sd{0.3} & 94.3\sd{0.3} \\
    FordA & 92.9\sd{0.4} & 92.9\sd{0.4} & 92.6\sd{0.4} & 92.6\sd{0.4} \\
    FordB & 81.9\sd{0.2} & 81.9\sd{0.2} & 79.5\sd{0.8} & 79.5\sd{0.8} \\
    Wafer & 99.7\sd{0.1} & 99.2\sd{0.2} & 99.5\sd{0.1} & 98.8\sd{0.2} \\
    \bottomrule
  \end{tblr}
\end{table}

\subsection{Short-Run Architectural Ablation (HAR)}
\label{app:supervised:har_ablation}

To make the architectural comparison visible while keeping training time modest, we include a 50-epoch ablation on HAR that contrasts \METHODNAME with a vanilla Transformer and two windowed variants without the full PMA mechanism.
For completeness we also include the 5\% SSL condition (same data, identical backbone depth/width; SSL losses only in that column)---see \Cref{tab:har_ablation_short}.

\begin{table}[tb]
  \centering
  \footnotesize
  \caption{\textbf{HAR (50 epochs).} Supervised vs.\ 5\% SSL. Top-1 / macro-F1 (\%); mean and standard deviation over three seeds.}
  \label{tab:har_ablation_short}
  \begin{tblr}{
    colspec={l*{4}{r}}
  }
    \toprule
    \SetCell[r=2]{l,valign=f} Model & \SetCell[c=2]{c} Supervised (CE) && \SetCell[c=2]{c} SSL (5\% labels, SVM probe) & \\
    \cmidrule[lr]{2-3}\cmidrule[lr]{4-5}
    & Top-1 & mF1 & Top-1 & mF1 \\
    \midrule
    \METHODNAME & 94.5\sd{0.1} & 94.6\sd{0.1} & 91.0\sd{2.0} & 91.1\sd{1.9} \\
    Vanilla Transformer & 92.4\sd{0.3} & 92.5\sd{0.2} & 85.6\sd{1.9} & 85.2\sd{2.1} \\
    Windowed Transformer (with temporal state pass) & 93.9\sd{0.6} & 94.0\sd{0.6} & 90.2\sd{1.2} & 90.2\sd{1.2} \\
    Windowed Transformer (no state pass) & 94.2\sd{0.5} & 94.3\sd{0.5} & 89.1\sd{0.8} & 89.2\sd{0.8} \\
    \bottomrule
  \end{tblr}
\end{table}

\paragraph{Takeaways.}
(i) In the short-run supervised setting, \METHODNAME is above the vanilla Transformer and on par with the windowed variants.
\;
(ii) Under SSL, \METHODNAME yields the strongest features at 5\% labels, suggesting the writable memory and progressive aggregation surface useful mid-range cues even when labels are scarce.

\subsection{Robustness to Reset-State Initialization}\label{app:supervised:reset}
The memory reset token \(M_r\) initializes the first-window state and can be mixed in by the learned reset gate when regimes change.
We tested robustness to the random initialization of \(M_r\) by repeating HAR training five times with different seeds; Table~\ref{tab:reset_robustness} shows a small spread ($\pm0.5$pp).

\begin{table}[tb]
  \centering
  \footnotesize
  \caption{\textbf{Reset-state robustness (HAR).} Supervised training (250 epochs, global seed 42) with different random seeds for \(M_r\); Top-1 accuracy (\%) on the official test split.}
  \label{tab:reset_robustness}
  \begin{tblr}{
    colspec={l*{5}{r}}
  }
    \toprule
    Seed 42 & Seed 123 & Seed 456 & Seed 789 & Seed 2024 & Mean $\pm$ Std \\
    \midrule
    93.69 & 94.30 & 94.91 & 93.96 & 94.71 & $94.31 \pm 0.51$ \\
    \bottomrule
  \end{tblr}
\end{table}

The spread is comparable to run-to-run variation, indicating that \METHODNAME is insensitive to the particular initialization of the reset memory.

\subsection{PMA Window-Size Ablation}\label{app:window_ablation}

\paragraph{Setup.} We study the sensitivity of \METHODNAME\ to the \emph{window length} $W$ in Progressive Memory Attention (PMA).
We parameterize $W$ as a fraction $\rho$ of the token length $K$ (after patchification), i.e., $W=\operatorname{round}(\rho K)$ with $\rho\in\{0.10, 0.20, 0.30, 0.50, 1.00\}$.
Unless noted, we keep the stride $S=\lfloor 0.5\,W\rfloor$, use a single memory slot per window ($n_m{=}1$), and leave every non-geometric hyperparameter unchanged (optimizer, temperatures for ICL/PCL/HGCL, Gaussian widths, augmentations).
We pretrain on \textsc{HAR} with the same recipe as in the main experiments but in a short-run setting (150 epochs); we then train SVM probes on the $1\%$ and $5\%$ label splits and report the average Top-1 accuracy and macro-F1.
Results are averaged across the same random seeds used for our ablations.

\begin{table}[tb]
  \centering
  \footnotesize
  \caption{\textbf{PMA window-size ablation on \textsc{HAR}} (150 epochs, three seeds). $W=\operatorname{round}(\rho K)$, $S=\lfloor W/2\rfloor$, $n_m{=}1$. We report the average over $1\%$ and $5\%$ label splits (mean and standard deviation over seeds). Performance peaks at $\rho=0.2$ and degrades as $W$ approaches full context.}
  \label{tab:pma-window-ablation}
  \begin{tblr}{
    colspec={lrr}
  }
    \toprule
    {Window fraction} $\rho$ & {Avg.\ Top-1 (\%)} & {Avg.\ mF1 (\%)} \\
    \midrule
    0.10 & 91.9\sd{1.1} & 91.9\sd{1.2} \\
    0.20 & 92.9\sd{0.9} & 93.0\sd{0.9} \\
    0.30 & 92.7\sd{0.7} & 92.8\sd{0.7} \\
    0.50 & 92.6\sd{0.7} & 92.7\sd{0.7} \\
    1.00 & 92.1\sd{0.4} & 92.2\sd{0.4} \\
    \bottomrule
  \end{tblr}
\end{table}

(1)~\textbf{Broad optimum at small/mid windows.} Performance varies by at most $0.3$pp for $\rho\!\in\![0.2,0.5]$, indicating that PMA's writable memory compensates for smaller windows by accumulating context progressively across windows and depth; this aligns with the receptive-field growth in \Cref{app:rfield}, \Cref{eq:hor-mem}.
(2)~\textbf{Very large windows are unnecessary.} As $\rho\!\to\!1.0$, accuracy and macro-F1 decline slightly.
With few very large windows, overlap reduces and the progressive mechanism has fewer horizontal updates, dampening the benefits of memory refresh.
(3)~\textbf{Practical choice.} Any $\rho$ in $[0.2,0.5]$ is a safe default.
We use $\rho{=}0.2$ in most of our configs to balance throughput (smaller attention tiles) and robust downstream accuracy without additional tuning.
For this ablation we did \emph{not} retune contrastive temperatures or Gaussian widths; the response in $[0.2,0.5]$ therefore represents a conservative estimate.
A separate screen over $n_m\in\{1,2,3,4\}$ on all eight datasets (150 epochs, three seeds) changes the average Top-1 by at most $0.6$pp ($85.2$, $84.8$, $85.3$, $85.1$); the range per dataset is at most $1.4$pp, except $2.6$pp on StarLightCurves.

\subsection{Losses Evaluation and Ablation}
\label{app:loss-ablation}

\paragraph{Per-dataset weights in the main experiments.}
The loss weights $\lambda_\text{ICL}$, $\lambda_\text{HGCL}$, and $\lambda_\text{PCL}$ are dataset-level hyperparameters in our main classification and cue-retention results.
They are tuned per dataset using the same sweep ranges reported in the released configurations and are not held at a uniform value across datasets.
The $\lambda{=}1$ uniform configuration reported in \Cref{tab:ablation-main-har} is a methodological reference point used to characterize the loss combination, not the configuration used to produce the headline numbers in \Cref{tab:classification_table}.

\paragraph{What the evaluations vary.}
Each evaluation in \Cref{tab:ablation-main} sweeps a single weight while holding the other two at $1$.
Specifically, \Cref{tab:lambda_sweeps} varies each $\lambda$ of the specific loss (\ie, from PCL, HGCL, and ICL) over $\{0, 0.01, 0.1, 0.5, 1\}$ on FordA, HAR, POC, and Wafer with the other two fixed to $1$;
\Cref{tab:ablation-main-har} reports leave-one-out and single-only configurations on top of the $\lambda{=}1$ reference point.
 Holding the remainders at $1$ makes the $\lambda{=}0$ entries of \Cref{tab:lambda_sweeps} coincide with the leave-one-out rows of \Cref{tab:ablation-main-har}, so both panels share one reference point.

\paragraph{Why the ${\sim}1$pp sweep stability statement holds.}
The conclusion that uniform weighting is a strong default (accuracy varies by under $0.4$pp on $\lambda_\text{PCL}\!\in\![0, 1]$ and under $1$pp on $\lambda_\text{HGCL}\!\in\![0, 1]$) is read off the sweep tables, \Cref{tab:lambda_sweeps}, where the remainders are held at $1$.
Practically, this means: if a practitioner deploys the protocol with all three losses at $\lambda{=}1$, the result was within $0.4$pp of a tuned configuration on these datasets; per-dataset tuning improves over uniform weighting but is not required.

\paragraph{Dataset selection for evaluation and ablation.}
The combined-loss ablation in \Cref{tab:ablation-main-har} and the per-weight sweeps in \Cref{tab:lambda_sweeps} both cover FordA, HAR, POC, and Wafer, with three pretraining seeds each.
Per-dataset tuning, where used in main results, follows the standard configuration sweep over a wider hyperparameter set and is not specific to the ablation tables. This has a practical implication for adopting the protocol on new datasets: the sweeps suggest that per-dataset $\lambda$ tuning produces measurable but small gains, and that the dominant choice---whether to deploy the multi-scale objective at all---is unlikely to depend on it.
Across the four datasets in \Cref{tab:ablation-main}, uniform $\lambda{=}1$ landed within $0.4$pp of tuned configurations, suggesting that practitioners adopting the protocol on new datasets can reasonably start from uniform weighting, with the larger gains coming from the protocol itself rather than from weight selection.

\FloatBarrier
\subsection{Additional Results for Learned Representations}
\label{app:repr}

\begin{figure}[tb]
\centering
\includegraphics{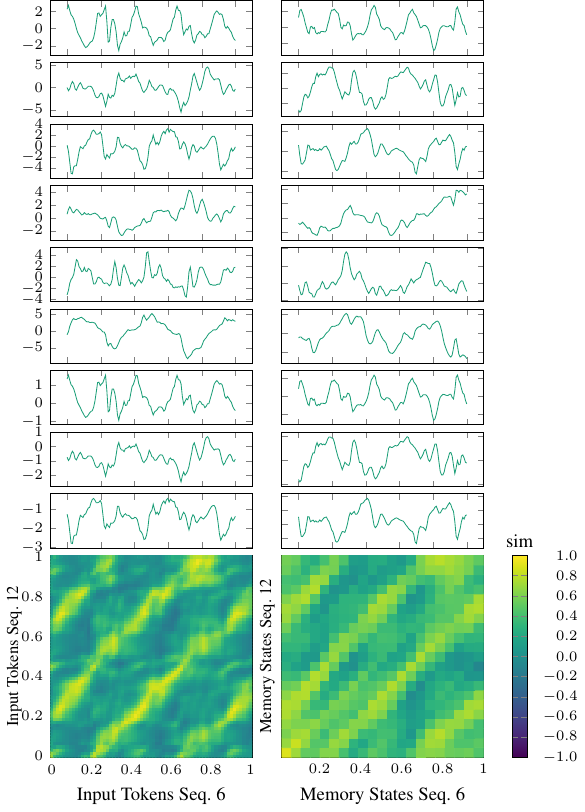}
\caption{The cosine similarity matrix between the input token (bottom left) and the memory states (bottom right) representations of two recordings (number 6 and 12, respectively, from the HAR validation set) of the same class (walking upstairs). The nine channels of the signals are plotted independently (top). Both matrices show off-center diagonal lines where the two recordings are in the same gait phase; the memory states repeat them at coarser resolution.}
\label{fig:har-same-class}
\end{figure}

\Cref{fig:har-same-class} shows two whole recordings of the same class (walking upstairs).
It shows four off-center diagonal lines, the time shifts at which the two recordings are in the same gait phase.
We included this supplementary figure for completeness, here plotting all 9 channels, rather than the single channel we plot for the other figures.
Additionally, this same-class comparison highlights similar patterns to what we see across all walking-related classes: the gait-cycle pattern (diagonal lines).

\begin{figure}[tb]
\centering
\resizebox{\linewidth}{!}{\includegraphics{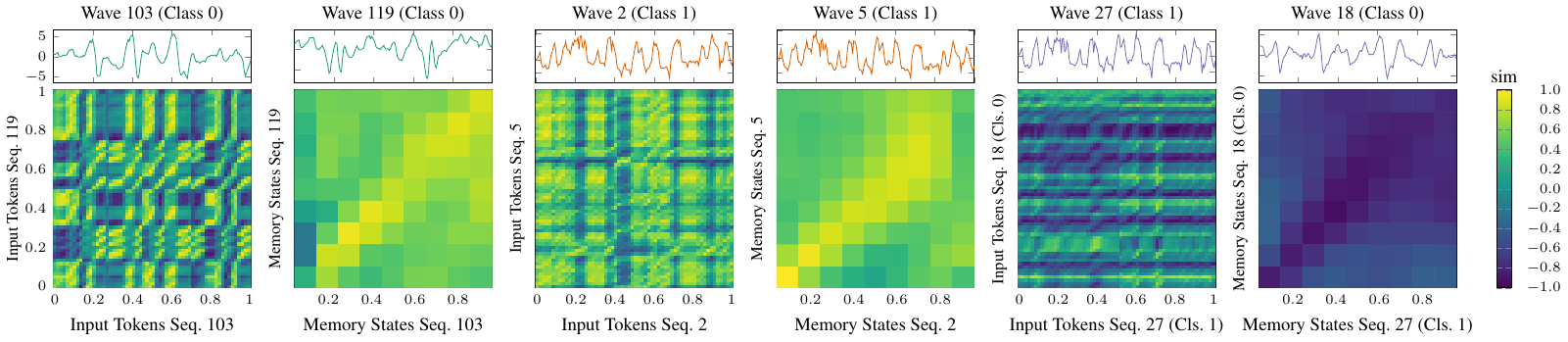}}
\caption{Cosine similarity matrices of the representations and states for the Epilepsy dataset.  Comparison of two sequences from the same class 0 (left) and class 1 (middle), and between two sequences from different classes (right).}
\label{fig:sim-dissim}
\end{figure}

\Cref{fig:sim-dissim} shows three sets of cosine similarity matrices from the Epilepsy dataset.
Each pair consists of a token representation comparison on the left and a memory state comparison on the right. The first column of each pair compares \emph{token embeddings}; the right column compares the corresponding \emph{progressive-memory states}. \textbf{Same-class pairs (columns 1--4):} Diagonal stripes are faint at token level, while the memory states of the two sequences are strongly similar. \textbf{Cross-class pair (columns 5--6):} The memory states of the two sequences are strongly dissimilar, much more so than their tokens. Overall, class separability is mainly realized \emph{after} recurrent aggregation; token embeddings alone retain limited spectral overlap.

\begin{figure}[tb]
\centering
\resizebox{\linewidth}{!}{\includegraphics{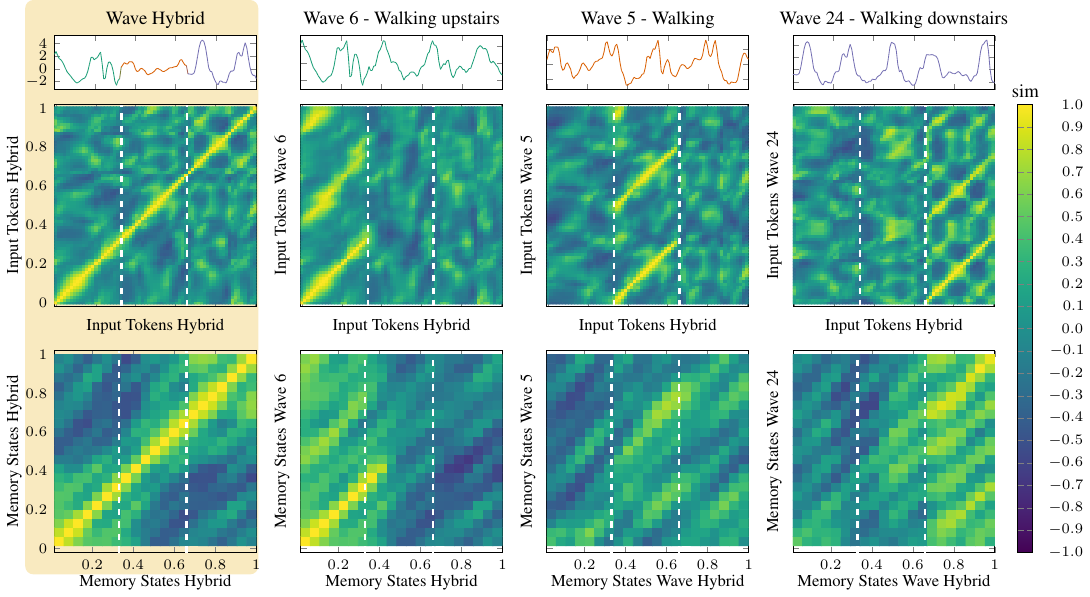}}
\caption{Cosine similarity matrices for the representations and states between pair-wise signals from HAR\@. Higher similarity shows that the signals correlate as evidenced by the learned embeddings. The vertical bars denote the different sections of the hybrid wave. The wave number corresponds to the sample index in the validation dataset.}
\label{fig:hybrid-sims-diff-classes}
\end{figure}

\Cref{fig:hybrid-sims-diff-classes} shows a 3-way splice of semantically close but unique class waveforms (walking upstairs, walking, and walking downstairs).
What we want to see in this figure is no correlation across splice borders.
The figure shows no such correlation at either the token or the memory-state level.

\begin{figure}[tb]
\centering
\resizebox{\linewidth}{!}{\includegraphics{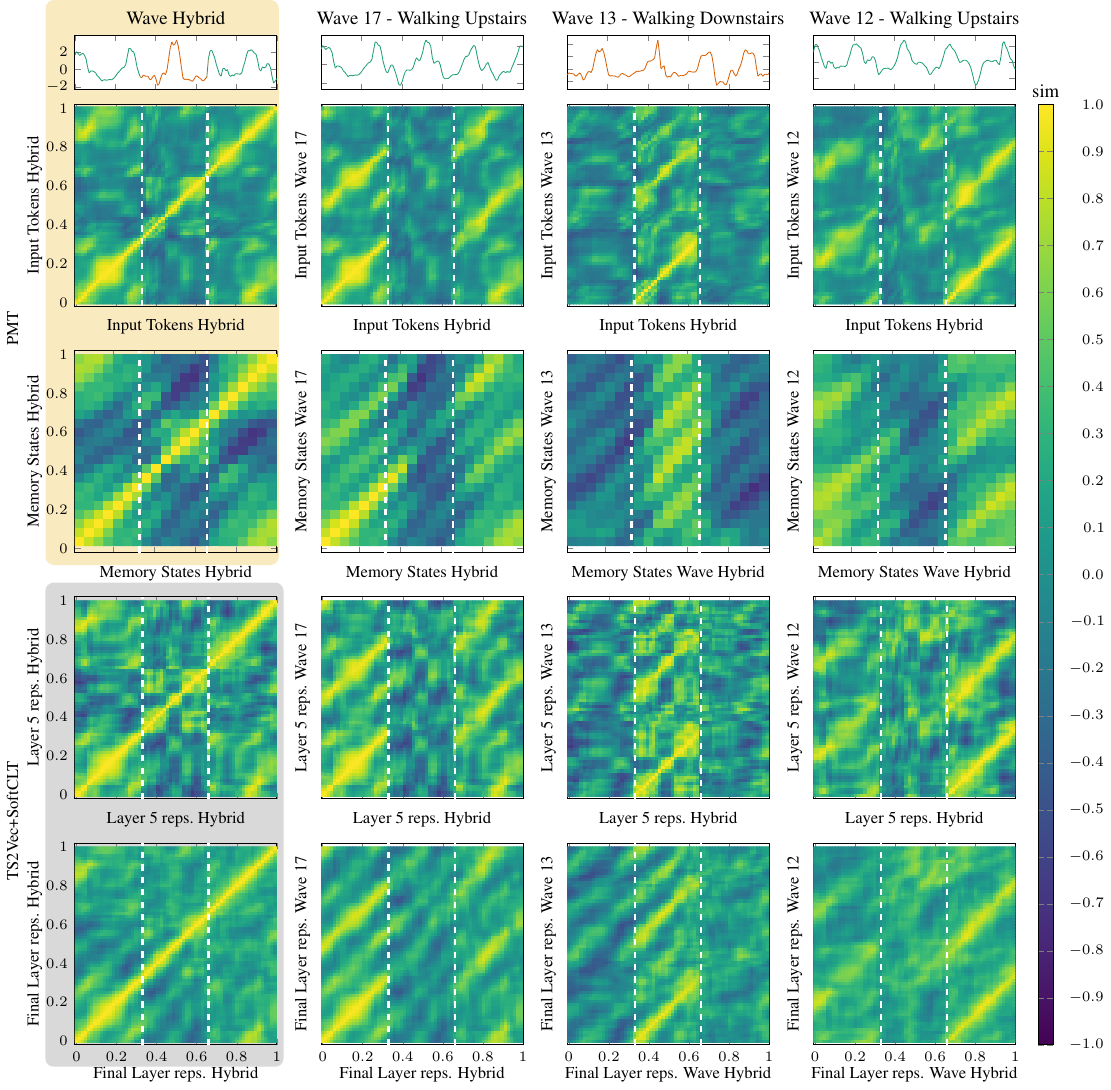}}
\caption{Cosine similarity matrices for the embeddings of \METHODNAME and TS2Vec + SoftCLT, from layer 5 and 10 (final), between pair-wise signals from HAR\@.  Higher similarity shows that the signals correlate as evidenced by the learned embeddings. The vertical bars denote the different sections of the hybrid wave. The wave number corresponds to the sample index in the validation dataset. Same samples as \Cref{fig:hybrid-sims}.}
\label{fig:softclt_fig4}
\end{figure}

\Cref{fig:softclt_fig4} shows TS2Vec+SoftCLT's embeddings at layer 5 and the final layer, from a TS2Vec encoder trained with the SoftCLT objective~\citep{lee2024soft}.
This figure uses the same data splice as in \Cref{fig:hybrid-sims}.
TS2Vec's final layer extracts the gait-cycle pattern we see in \Cref{fig:hybrid-sims}, but no ``checkerboard'' pattern in the hybrid-to-hybrid comparison; its intermediate layer~5 shows a weaker one than \METHODNAME's memory states.
We include this figure for completeness; although extracting representations from different layers may not be perfectly analogous to our token vs.\ memory-state comparison, it illustrates the difference in mid-range motif capture.

To further showcase the behavior of the representations at different structural levels, we visualize t-SNE and PCA projections of two HAR splices.
The t-SNE projection uses the splice of \Cref{fig:hybrid-sims}; both illustrate how the representation space evolves from fine-grained tokens to compressed memory states.

\begin{figure}[tb]
\centering
\resizebox{\linewidth}{!}{\includegraphics{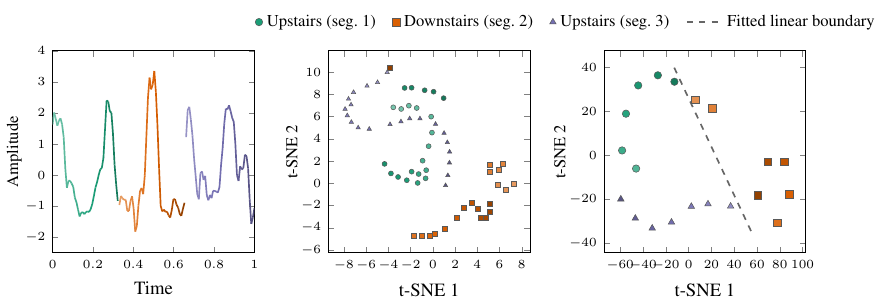}}
\caption{(Left) Double spliced HAR waveform. (Middle) t-SNE of representations (tokenizer tokens) of the signal. (Right) t-SNE of memory-states. The splice joins three validation recordings: Upstairs, Downstairs and a second Upstairs recording. Tokens and memory states are shown after removing the mean representation at each position over the validation set. Colour marks the segment and shade the position within it (light = start); the dashed line is a linear boundary fitted to the memory states.}
\label{fig:pca-repr-standing}
\end{figure}

\begin{figure}[!ht]
\centering
\resizebox{\linewidth}{!}{\includegraphics{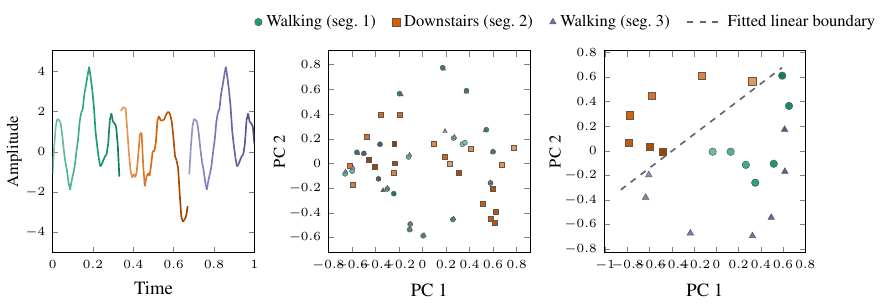}}
\caption{(Left) Double spliced HAR waveform. (Middle) PCA of representations (tokenizer tokens) of the signal. (Right) PCA of memory-states. The splice joins two validation recordings: Walking, Walking Downstairs and the same Walking recording again. Tokens and memory states are shown after removing the mean representation at each position over the validation set. Colour marks the segment and shade the position within it (light = start); the dashed line is a linear boundary fitted to the memory states.}
\label{fig:pca-repr}
\end{figure}

First, \Cref{fig:pca-repr-standing} projects the representations of an Upstairs--Downstairs--Upstairs splice of three recordings using t-SNE; the memory states of each segment occupy their own region.
Similarly, \Cref{fig:pca-repr} shows a PCA projection over another unique HAR splice alternating between walking and walking downstairs.
Because the PMA unrolls over sliding windows with a stride greater than one, the model produces a larger number of token representations (capturing \emph{local nuance}) compared to the more compressed memory states (capturing \emph{mid-range motifs}).
The tokens of the two activities mix (center panel), while the memory states of each segment occupy their own region (right panel); the states of the repeated Walking segment start beside the Downstairs states and move back towards those of the first segment, carrying the preceding context (however, we note that the first memory state in the sequence is typically less semantically distinct due to its small initial receptive field, as derived in \Cref{app:rfield}).

\section{Limitations}
\label{app:limitations}

\METHODNAME's framework is independent of the input stem we use, but in this work we tokenize via a fixed-stride 1D convolution, the standard choice for patch-based time-series architectures. As \Cref{sec:patch-and-stride-ab} shows, patch length is well-behaved across a wide plateau but stride must be small enough to capture short-lived events, making tokenization the main per-dataset adjustment in our recipe. Replacing the patch stem with a learned or signal-adaptive tokenizer is a natural avenue for removing this tuning step.

The three contrastive objectives, particularly ICL at the sequence level, depend on large in-batch negative sets. This is a standard characteristic of InfoNCE-based self-supervised frameworks but constrains application to extremely small datasets or resource-limited hardware without falling back on momentum queues, which we did not explore.

\section{LLM Disclosure}
We used a large language model as a writing and editing assistant, and to help surface potentially relevant related work during scoping.
The authors take responsibility for all technical claims, modeling choices, experiments, and analyses.
No citations were included without verification.
No empirical results were generated with LLMs.

\clearpage
\section*{NeurIPS Paper Checklist}

\begin{enumerate}

\item {\bf Claims}
    \item[] Question: Do the main claims made in the abstract and introduction accurately reflect the paper's contributions and scope?
    \item[] Answer: \answerYes{}
    \item[] Justification: The claims regarding the multi-scale contrastive learning framework and Progressive Memory Transformer (PMT) are directly supported by our evaluations on downstream tasks (global classification, mid-range cue retention, and local forecasting) in \Cref{sec:experiments}.

\item {\bf Limitations}
    \item[] Question: Does the paper discuss the limitations of the work performed by the authors?
    \item[] Answer: \answerYes{}
    \item[] Justification: We explicitly discuss limitations regarding tokenizer sensitivity (patch size and stride) and the requirement of large batch sizes for contrastive losses in \Cref{app:limitations}.

\item {\bf Theory assumptions and proofs}
    \item[] Question: For each theoretical result, does the paper provide the full set of assumptions and a complete (and correct) proof?
    \item[] Answer: \answerNo{}
    \item[] Justification: \Cref{app:rfield} states the receptive-field bounds as formal claims with their assumptions; they follow directly from the window geometry, and we do not give full proofs.

    \item {\bf Experimental result reproducibility}
    \item[] Question: Does the paper fully disclose all the information needed to reproduce the main experimental results of the paper to the extent that it affects the main claims and/or conclusions of the paper (regardless of whether the code and data are provided or not)?
    \item[] Answer: \answerYes{}
    \item[] Justification: We detail the experimental protocols, datasets, and hyperparameters in \Cref{sec:experiments} and \Cref{app:implementation,app:protocol,app:additional}, and provide access to the full source code and configurations.

\item {\bf Open access to data and code}
    \item[] Question: Does the paper provide open access to the data and code, with sufficient instructions to faithfully reproduce the main experimental results, as described in supplemental material?
    \item[] Answer: \answerYes{}
    \item[] Justification: The source code and instructions to reproduce the results are provided via the public repository linked on the first page of the main paper.

\item {\bf Experimental setting/details}
    \item[] Question: Does the paper specify all the training and test details (e.g., data splits, hyperparameters, how they were chosen, type of optimizer) necessary to understand the results?
    \item[] Answer: \answerYes{}
    \item[] Justification: Training details, including optimizers, learning rate schedules, and batch sizes, are specified in \Cref{app:implementation}, with dataset and evaluation protocol details in \Cref{app:protocol,app:additional}.

\item {\bf Experiment statistical significance}
    \item[] Question: Does the paper report error bars suitably and correctly defined or other appropriate information about the statistical significance of the experiments?
    \item[] Answer: \answerYes{}
    \item[] Justification: \Cref{tab:classification_table,tab:forda-cue-eos,tab:forecasting,tab:xlstm-comparison,tab:ablation-main,tab:ablation-patch-stride}, as well as \Cref{tab:supervised_full,tab:har_ablation_short,tab:reset_robustness,tab:pma-window-ablation}, report means with sample standard deviations over independent seeds (ten for \Cref{tab:classification_table,tab:xlstm-comparison}, five for \Cref{tab:forda-cue-eos,tab:forecasting,tab:reset_robustness}, three otherwise).

\item {\bf Experiments compute resources}
    \item[] Question: For each experiment, does the paper provide sufficient information on the computer resources (type of compute workers, memory, time of execution) needed to reproduce the experiments?
    \item[] Answer: \answerYes{}
    \item[] Justification: Detailed computational resource information, including GPU type (NVIDIA GH200), memory bounds, and execution times, is provided in \Cref{app:compute}, with per-architecture training and inference cost in \Cref{tab:arch-cost}.

\item {\bf Code of ethics}
    \item[] Question: Does the research conducted in the paper conform, in every respect, with the NeurIPS Code of Ethics \url{https://neurips.cc/public/EthicsGuidelines}?
    \item[] Answer: \answerYes{}
    \item[] Justification: The research complies with the NeurIPS Code of Ethics and does not involve harmful applications or sensitive human data without consent.

\item {\bf Broader impacts}
    \item[] Question: Does the paper discuss both potential positive societal impacts and negative societal impacts of the work performed?
    \item[] Answer: \answerNA{}
    \item[] Justification: This paper focuses on foundational architectural improvements for time-series modeling and does not directly tie to specific applications that introduce immediate or novel negative societal impact concerns.

\item {\bf Safeguards}
    \item[] Question: Does the paper describe safeguards that have been put in place for responsible release of data or models that have a high risk for misuse (e.g., pre-trained language models, image generators, or scraped datasets)?
    \item[] Answer: \answerNA{}
    \item[] Justification: The proposed foundational time-series method and models do not pose a high risk for misuse, and we are not releasing scraped datasets or highly sensitive generative systems.

\item {\bf Licenses for existing assets}
    \item[] Question: Are the creators or original owners of assets (e.g., code, data, models), used in the paper, properly credited and are the license and terms of use explicitly mentioned and properly respected?
    \item[] Answer: \answerYes{}
    \item[] Justification: All benchmark datasets and baseline methods used in this work are cited and publicly available, and we follow the terms of use set by their providers.

\item {\bf New assets}
    \item[] Question: Are new assets introduced in the paper well documented and is the documentation provided alongside the assets?
    \item[] Answer: \answerYes{}
    \item[] Justification: The source code and configurations are released in the public repository linked on the first page, with documentation alongside them.

\item {\bf Crowdsourcing and research with human subjects}
    \item[] Question: For crowdsourcing experiments and research with human subjects, does the paper include the full text of instructions given to participants and screenshots, if applicable, as well as details about compensation (if any)?
    \item[] Answer: \answerNA{}
    \item[] Justification: The paper does not involve crowdsourcing or new research with human subjects.

\item {\bf Institutional review board (IRB) approvals or equivalent for research with human subjects}
    \item[] Question: Does the paper describe potential risks incurred by study participants, whether such risks were disclosed to the subjects, and whether Institutional Review Board (IRB) approvals (or an equivalent approval/review based on the requirements of your country or institution) were obtained?
    \item[] Answer: \answerNA{}
    \item[] Justification: No human subjects were involved in the conduction of this research.

\item {\bf Declaration of LLM usage}
    \item[] Question: Does the paper describe the usage of LLMs if it is an important, original, or non-standard component of the core methods in this research? Note that if the LLM is used only for writing, editing, or formatting purposes and does \emph{not} impact the core methodology, scientific rigor, or originality of the research, declaration is not required.
    \item[] Answer: \answerNA{}
    \item[] Justification: LLMs were not used as a component of the core methodology. As disclosed in the final section of the Appendix, they were used as a writing and editing assistant.

\end{enumerate}

\end{document}